\documentclass[lettersize,journal]{IEEEtran}
\usepackage{amsmath,amsfonts}
\usepackage{algorithmic}
\usepackage{algorithm}
\usepackage{array}
\usepackage[caption=false,font=normalsize,labelfont=sf,textfont=sf]{subfig}
\usepackage{textcomp}
\usepackage{stfloats}
\usepackage{url}
\usepackage{verbatim}
\usepackage{graphicx}
\usepackage{hyperref}
\usepackage{indentfirst}
\usepackage{bm}
\usepackage{multirow}
\usepackage{multicol}
\usepackage{makecell}
\usepackage{booktabs}
\usepackage{bm}
\usepackage{bbm}
 \usepackage{pifont}
\usepackage{subfloat}
\usepackage{xcolor}

\newcommand{\Tref}[1]{Tab.~\ref{#1}}

\newcommand{\Fref}[1]{Fig.~\ref{#1}}
\newcommand{\Sref}[1]{Section~\ref{#1}}
\newcommand{\revise}[1]{\textcolor{black}{#1}}

\begin{document}

\title{Siamese-DETR for Generic Multi-Object Tracking}


\author{Qiankun Liu, Yichen Li, Yuqi Jiang, Ying Fu,~\IEEEmembership{Senior Member,~IEEE}
\thanks{Qiankun Liu, Yichen Li, Yuqi Jiang and Ying Fu are with School of Computer Science and Technology, Beijing Institute of Technology; Email: \{liuqk3, liyichen, yqjiang, fuying\}@bit.edu.cn; Ying Fu is the corresponding author.}
\thanks{This paper has supplementary downloadable material available at http://ieeexplore.ieee.org, provided by the author. }
}

\markboth{Journal of \LaTeX\ Class Files,~Vol.~14, No.~8, August~2021}%
{Shell \MakeLowercase{\textit{et al.}}: A Sample Article Using IEEEtran.cls for IEEE Journals}

\maketitle

\begin{abstract}
  The ability to detect and track the dynamic objects in different scenes is fundamental to real-world applications, \textit{e.g.}, autonomous driving and robot navigation. However, traditional Multi-Object Tracking (MOT) is limited to \revise{track} objects belonging to the pre-defined closed-set categories. Recently, \revise{Generic MOT (GMOT) is proposed to track interested objects beyond pre-defined categories and it can be divided into Open-Vocabulary MOT (OVMOT) and Template-Image-based MOT (TIMOT). Taking the consideration that the expensive well pre-trained (vision-)language model and fine-grained category annotations are required to train OVMOT models, in this paper,} we focus on \revise{TIMOT} and propose a simple but effective method, Siamese-DETR. Only the commonly used detection datasets (\emph{e.g.}, COCO) are required for training. Different from existing \revise{TIMOT} methods, which train a Single Object Tracking (SOT) based detector to detect interested objects and then apply a data association based MOT tracker to get the trajectories, we leverage the inherent object queries in DETR variants. Specifically: 1) The multi-scale object queries are designed based on the given template image, which are effective for detecting different scales of objects with the same category as the template image; 2) A dynamic matching training strategy is introduced to train Siamese-DETR on commonly used detection datasets, which takes full advantage of provided annotations; 3) The online tracking pipeline is simplified through a tracking-by-query manner by incorporating the tracked boxes in the previous frame as additional query boxes. The complex data association is replaced with the much simpler Non-Maximum Suppression (NMS). Extensive experimental results show that Siamese-DETR surpasses existing MOT methods on GMOT-40 dataset by a large margin. Codes are avaliable at \textcolor{blue}{\url{https://github.com/yumu-173/Siamese-DETR}}.

\end{abstract}


\section{Introduction}

\IEEEPARstart{M}ulti-Object Tracking (MOT) aims at estimating the locations of interested objects in the given video while maintaining their identities consistently, which has various applications, such as autonomous driving, robot navigation, video surveillance, and so on. Benefiting from the advances in object detection, the tracking-by-detection paradigm has become popular for MOT in the past decade. Though great success has been made, the generalization ability of existing MOT methods still needs to be improved due to the limited pre-defined closed-set categories, like pedestrian~\cite{bergmann2019tracking, braso2020learning, wojke2017simple, chu2019famnet, zhang2022bytetrack}, car~\cite{zhou2020tracking}, \emph{etc}. 

To overcome the aforementioned drawback of traditional MOT task, \revise{Generic Multi-Object (GMOT) is} recently introduced and tries to track objects of arbitrary categories. \revise{It is based on the assumption that at test time we are given the descriptions of interested objects. According to the types of descriptions, GMOT can be divided into Open-Vocabulary MOT (OVMOT)~\cite{li2023ovtrack} and Template-Image-based MOT (TIMOT)~\cite{bai2021gmot} tasks. Among them,} OVMOT methods use the text prompt (\textit{e.g.}, category name) as the description, while \revise{TIMOT} methods utilize the template image as the description. Both types of descriptions are flexible and enlarge the closed-set categories to an open-set one, making multi-object tracking methods more suitable for real-world applications. However, due to the domain gap between text and image, the well pre-trained (vision-)language models (\textit{e.g.}, BERT~\cite{devlin2018bert} and CLIP~\cite{radford2021learning}) and fine-grained category annotations are needed to train the detectors in OVMOT methods. Except that annotating fine-grained category information is laborious and professional, \revise{the pre-training of the (vision-)language model} requires a huge amount of training data and computational resources, making the utilization of OVMOT methods expensive. Taking this into consideration, we focus on \revise{TIMOT} in this paper and propose a simple but effective method, Siamese-DETR. Only the commonly used detection datasets (\textit{e.g.}, COCO~\cite{lin2014microsoft}) are required to train the proposed method.

\begin{figure}[t]
	\centering
	\includegraphics[width=\linewidth]{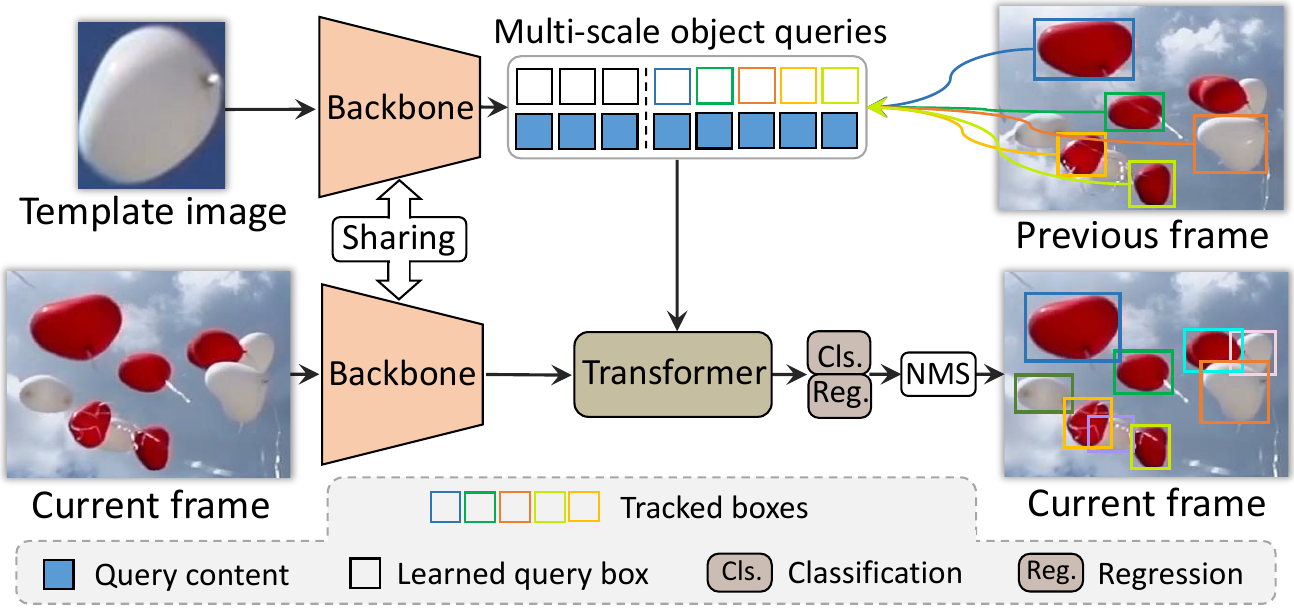}
	\caption{The online tracking pipeline of Siamese-DETR for generic multi-object tracking \revise{based on template image}. The template image is fed into the backbone network to get the query contents, while the query boxes consist of the learned query boxes and the tracked boxes in the previous frame. With this design, the objects in \revise{current} frame are tracked by their corresponding boxes, while the missed objects in the previous frame (but still exist in the current frame) or newly appeared objects in the current frame are detected and tracked by the learned query boxes.}
	\label{fig_motivation}
\end{figure}

Early \revise{TIMOT} methods~\cite{luo2013generic,luo2014bi} learn a Support Vector Machine (SVM) for each object identity through multiple task learning~\cite{evgeniou2004regularized} based on hand-crafted features (\emph{e.g.}, HoG~\cite{dalal2005histograms}). The identities of different objects are involuntarily maintained since each of them is detected and tracked independently by their dedicated SVMs. Recently, inspired by the success of traditional MOT task, where the tracking-by-detection paradigm~\cite{luo2013generic,luo2014bi,bai2021gmot, fu2023category} dominates the mainstream and achieves appealing performance, the newly proposed \revise{TIMOT} method~\cite{bai2021gmot} also follows the tracking-by-detection paradigm. Specifically, the tracking pipeline is divided into object detection and object tracking stages: 1) For object detection, a Single Object Tracking (SOT)~\cite{huang2020globaltrack} based detector is designed to detect all the objects that share the same category with the template image. Since there is no provided training data for \revise{TIMOT} task~\cite{bai2021gmot}, SOT datasets (\textit{e.g.}, LaSOT~\cite{fan2021lasot} and GOT-10K~\cite{huang2019got}) and object detection dataset (\textit{e.g.}, COCO~\cite{lin2014microsoft}) are used for the training of detector; 2) For object tracking, existing MOT trackers (\emph{e.g.}, SORT~\cite{bewley2016simple}, DeepSORT~\cite{wojke2017simple}, IOU~\cite{bochinski2017high}, \textit{etc}) are directly utilized as data association algorithms to get the trajectories of different objects. Unfortunately, the tracking performance is still moderate even it is of high complexity. The \revise{TIMOT} task still needs to be well studied to achieve \revise{better overall tracking performance while simplifying the tracking pipeline}.

In this paper, we leverage the inherent object queries in DETR variants~\cite{zhang2022dino,zhu2020deformable,li2022dn,liu2022dab, li2024supervise} and propose a simple but effective method, Siamese-DETR. As shown in \Fref{fig_motivation}, the object queries contain the information of the template image for detection and the tracked boxes for tracking. Although Siamese-DETR follows the tracking-by-detection paradigm, the detection and tracking are performed simultaneously. The complex data association procedure is replaced by a much simpler Non-Maximum Suppression (NMS) to remove some duplicated boxes. Compared with existing methods, Siamese-DETR detects interested objects more effectively and tracks objects more simply.

To detect interested objects with the given template image effectively, the Multi-Scale Object Queries (MSOQ) and Dynamic Matching Training Strategy (DMTS) are designed: 1) Multi-scale object queries. The decoupled object query~\cite{liu2022dab} that consists of query content and query box is adopted, where the query content is obtained from the template image while the query box is learned during training. In detail, we feed the template image into the backbone network of the detector to get hierarchical multi-scale features and map each scale of them into a query content. The multi-scale query contents (\emph{e.g.}, 4 scales) are equally replicated to match with the number of learned query boxes (\emph{e.g.}, 600). Since the features with different scales are sensitive to objects of different scales, Siamese-DETR detects different scales of objects that share the same category with the template image effectively; 
2) Dynamic matching training strategy. Given a training image in commonly used detection datasets (\emph{e.g.}, COCO~\cite{lin2014microsoft}), the corresponding annotations are all utilized more than once. Specifically, the objects that share the same category with the template image are treated as positive samples while the others are treated as negative samples. The introduction of negative samples takes full advantage of the provided annotations. By sampling more than one template image for each training image, the annotations can be dynamically used more than once, which benefits Siamese-DETR further.

To track objects simply, we propose a Tracking-by-Query (TbQ) strategy. The tracked boxes are used as additional query boxes and the query denoising is optimized to adapt to \revise{TIMOT}: 1) Tracked boxes as additional query boxes. The tracked boxes in the previous frame are paired with the query contents to serve as additional object queries. Object queries with tracked boxes and learned boxes are responsible for tracking and detection respectively and independently. The simple NMS is utilized to remove the detected boxes that are duplicated with tracked boxes; 2) Optimized query denoising. Since there is no video training data for \revise{TIMOT}~\cite{bai2021gmot}, the query denoising~\cite{li2022dn} strategy is optimized from common object detection and adopted to mimic the tracking scenarios in static images. The experimental results demonstrate that Siamese-DETR surpasses existing MOT methods on GMOT-40~\cite{bai2021gmot} by a large margin. In summary, the contributions of this work are as follows:
\begin{itemize}
    \item We propose Siamese-DETR for \revise{template-image-based multi-object tracking} and introduce multi-scale object queries to effectively detect different scales of objects that share the same category with the template image. 
    \item We introduce a dynamic matching training strategy for Siamese-DETR, enabling the training on commonly used detection datasets effectively.
    \item We design a simple online tracking strategy by incorporating the tracked boxes as additional query boxes. Objects are tracked in a tracking-by-query manner. 
\end{itemize}

The remainder of this paper is organized as follows: \Sref{sec_related_work} firstly reviews the related works about object tracking and DETR variants. Next, the details of the proposed method are illustrated in \Sref{sec_method}. Then, we provide the implementation details and compare the proposed method with existing methods in \Sref{sec_experiments}. Finally, we provide the conclusion and the discussions on the limitations of the proposed method in \Sref{sec_conclusion}.

\section{Related Work}
\label{sec_related_work}
This section briefly reviews related works from different aspects, including multi-object tracking, \revise{template-image-based} multi-object tracking, open-vocabulary multi-object tracking and DETR variants. 

\subsection{Multi-Object Tracking} 

In the past decade, Multi-Object Tracking (MOT) has emerged as a popular research area and has been dominated by the tracking-by-detection paradigm. Existing tracking-by-detection methods involve object detection and data association stages, and can be divided into offline and online methods. Offline methods~\cite{roshan2012gmcp, zhang2008global, keuper2015efficient, tang2017multiple, chen2023instance, fu2023raw}  process the video in a batch way and even can utilize the whole video information to handle the data association problem better. Differently, online methods~\cite{bewley2016simple, fang2018recurrent, bergmann2019tracking, braso2020learning, wojke2017simple, chu2019famnet, zhang2022bytetrack,zhang2024deep} process the video frame-by-frame and generate trajectories only using information up to the current frame, which is more suitable for causal applications than offline ones. 

Traditional MOT methods mainly focus on data association problem, including Hungarian algorithm, network flow~\cite{roshan2012gmcp, zhang2008global,ren2020tracking}, and graph multicut~\cite{keuper2015efficient, tang2017multiple}. Among them, except the Hungarian algorithm, others can only be performed in an offline manner. In recent years, with the advancement of deep learning and object detection, online tracking has attracted more and more attention. On contrary to offline methods, online methods usually adopt the Hungarian algorithm for data association, but focus on the joint learning of object detection and some useful priors, such as object motions~\cite{bergmann2019tracking,zhou2020tracking,lee2023decode,fu2022low}, appearance features~\cite{chu2019famnet,liu2022online,wan2021tracking}, occlusion maps~\cite{liu2022online}, \revise{object relations~\cite{liu2020gsm}} and so on. However, except for the annotation of box and category ID, extra annotations are required for the learning of these priors, \emph{e.g.}, object identity for appearance feature learning.

Though great progress has been made in MOT, most existing methods are designed to track objects that are limited to a pre-defined small closed-set of categories. For example,  car and pedestrian. In this paper, we focus on generic multi-object tracking to extend the closed-set of categories in MOT to an open-set of generic categories, which are not limited to several specific ones.

\subsection{\revise{Template-Image-based Multi-Object Tracking}}
The \revise{Template-Image-based Multi-Object Tracking (TIMOT)} task is introduced to address the generalization issue in MOT about ten years ago~\cite{luo2013generic,luo2014bi}. Similar to MOT, \revise{TIMOT} follows the tracking-by-detection paradigm~\cite{luo2013generic,luo2014bi,bai2021gmot}. However, much less attention has been paid to \revise{TIMOT}, which is quite different from MOT. The main reason is that the data that is suitable for \revise{TIMOT} is scarce. Recently, GMOT-40~\cite{bai2021gmot} has been developed as a public dataset for the evaluation of \revise{TIMOT}. Nevertheless, no well-annotated training data is available for \revise{TIMOT}. 

Early methods~\cite{luo2013generic,luo2014bi} track generic multi-objects based on Support Vector Machine (SVM) and hand-crafted features. Each object is detected and tracked by a dedicated SVM. The SVM is initialized based on the given template image and updated in an online manner while tracking. Recently, the newly proposed method~\cite{bai2021gmot,li2023inference} firstly detects all objects that share the same category with the template image through a Single Object Tracking (SOT) based detector (specifically, GlobalTrack~\cite{huang2020globaltrack}), then some online data association MOT trackers (\emph{e.g.}, SORT~\cite{bewley2016simple}, DeepSORT~\cite{wojke2017simple}, IOU~\cite{bochinski2017high}, \textit{etc}) are applied to get the trajectories of objects. To mitigate the gap between SOT and \revise{TIMOT}, single object tracking datasets (LaSOT~\cite{fan2021lasot}, GOT-10K~\cite{huang2019got}) and object detection dataset (COCO~\cite{lin2014microsoft}) are used to train the SOT based detector. However, the tracking performance is far from satisfactory. 

Different from existing \revise{TIMOT} method~\cite{bai2021gmot} that detects objects based on SOT tracker~\cite{huang2020globaltrack}, we leverage the advantage of object queries in DETR variants for object detection and tracking. With the proper query design and training strategy, our method distinguishes the interested objects from others effectively. In addition, the tracking pipeline is also simplified by incorporating the tracked boxes into object queries.

\subsection{Open-Vocabulary Multi-Object Tracking}

With the recent development of language~\cite{devlin2018bert,vaswani2017attention} and vision-language models~\cite{radford2021learning}, the Open-Vocabulary Multi-Object Tracking (OVMOT)~\cite{li2023ovtrack} is proposed to track objects that belong to arbitrary categories. 

Similar to traditional MOT and \revise{TIMOT}, OVMOT also follows the tracking-by-detection pipeline, where open-vocabulary object detection plays a key role.  OVTrack~\cite{li2023ovtrack} localizes interested objects with a class agnostic R-CNN (\textit{i.e.}, Faster R-CNN~\cite{ren2015faster}) and the vision-language model (\textit{i.e.}, CLIP~\cite{radford2021learning}). Specifically, all objects, including the ones that are not interested, are detected by the R-CNN, where the object features are aligned with the counterparts extracted by the image encoder in CLIP through knowledge distillation. Then the interested objects are selected by comparing the similarity between object features and text features extracted by the text encoder in CLIP.  Finally, a data association procedure is adopted to link objects in adjacent frames. Similarly, GLIP~\cite{li2022grounded} detects objects using DyHead~\cite{dai2021dynamic} and BERT~\cite{devlin2018bert}. The text and image features are aligned with each other by iteratively fusing them in several successive blocks rather than supervising the model with knowledge distillation. Though OVMOT trackers can track arbitrary object categories, expensive well pre-trained language or vision-language models are required to handle the domain gap between texts and images. In addition, laborious fine-grained category annotations are also needed to help the model recognize accurate objects with different text descriptions. For example, OVTrack is trained on LVIS~\cite{gupta2019lvis} with 1200+ category annotations and GLIP is trained on Objects365~\cite{shao2019objects365} with 356 (which is further increased to 1300+ by the authors) category annotations. 

Compared with the aforementioned methods, the proposed Siamese-DETR does not require the expensive pre-trained (vision-)language model nor the laborious fine-grained category annotations. It achieves better performance when only the COCO~\cite{lin2014microsoft} dataset (with 80 categories) is used for training.

\subsection{DETR Variants} 

DETR~\cite{carion2020end} is the first end-to-end object detector. The main idea in it is the object query and the Hungarian loss. The anchor boxes and NMS components are abandoned, reducing the complexity of detectors significantly. However, DETR suffers from slow convergence. Lots of works are proposed to address this issue. 

Deformable-DETR~\cite{zhu2020deformable} replaces the common attention with deformable attention, which reduces the computational cost and \revise{makes} it possible to use multi-scale features for object detection. DAB-DETR~\cite{liu2022dab} decouples the object queries into learnable contents and learnable boxes. The query boxes are iteratively updated at each decoder layer. DN-DETR~\cite{li2022dn} finds that the slow convergence of DETR is mainly caused by the unstable matching between object queries and ground-truth boxes. To reduce the instability, DN-DETR introduces a denoising training approach to accelerate the convergence. Specifically, except for the object queries, noisy ground-truth boxes and labels are additionally fed into the decoder, which improves the model's ability of box regression and classification. Once the training procedure is finished, the object queries in the aforementioned DETR variants are fixed. All images share the same ones, which can not be dynamically updated according to the input images. To solve this, DINO~\cite{zhang2022dino} proposes a mixed query selection mechanism, where the query boxes are dynamically selected based on the image features.

In this paper, multi-scale object queries are designed, which contain the information of the template image. The tracked boxes are further used as additional query boxes. Object detection and tracking are performed simultaneously.

\begin{figure*}
    \centering
    \includegraphics[width=\linewidth]{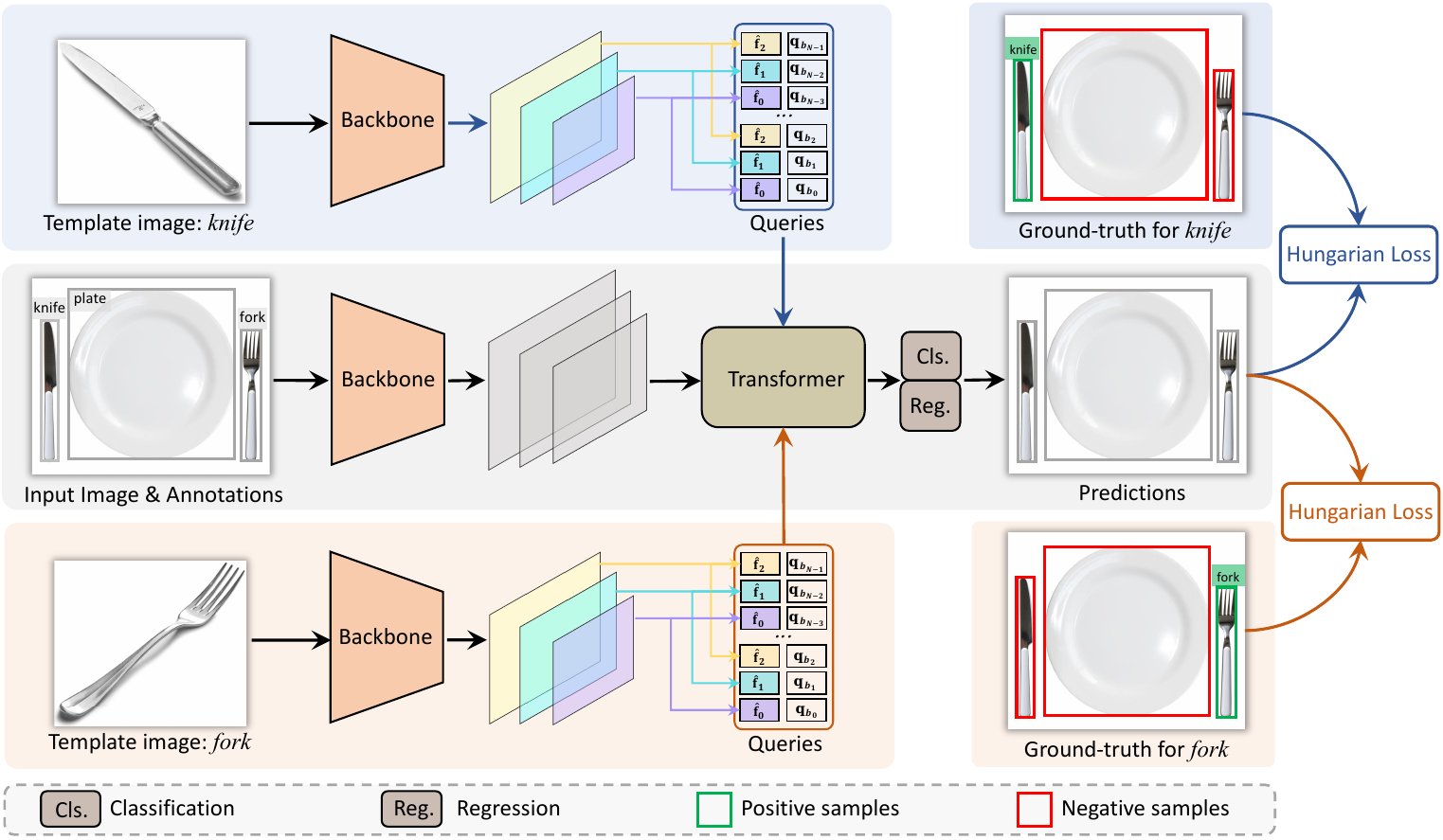}
    \caption{Overview of Siamese-DETR in the training stage. The multi-scale object queries are decoupled into learnable query boxes and query contents. The query contents are mapped from the multi-scale features extracted from the template image by the backbone network. The model is trained with Hungarian loss~\cite{carion2020end} and the proposed dynamic matching training strategy which turns the provided annotations into positive and negative samples dynamically according to the given template image. For simplicity, the optimized query denoising is not presented in the figure.}
    \label{fig_training_stage}
\end{figure*}

\section{Methodology}
\label{sec_method}
In this section, we first present the overall architecture of Siamese-DETR. Then, we introduce the multi-scale object queries for the detection of objects that share the same category with the template image. Next, we introduce the dynamic matching training strategy that trains Siamese-DETR on commonly used detection datasets. \revise{Furthermore}, we show how to apply Siamese-DETR to online tracking straightforwardly and simply in a tracking-by-query manner. \revise{Finally, the training details of Siamese-DETR are presented.}

\subsection{Overview}

The overview of the proposed Siamese-DETR in the training stage is shown in \Fref{fig_training_stage}. Siamese-DETR contains a backbone network \revise{(\textit{e.g.}, Swin Transformer~\cite{liu2021swin})}, a transformer (including the encoder and the decoder)~\cite{vaswani2017attention, liu2024transformer}, a detection head for classification and box regression \revise{(the same with DINO~\cite{zhang2022dino})}, and a set of object queries. In order to detect objects of different scales that share the same category with the template image, the Multi-Scale Object Queries (MSOQ, \Sref{sec_multi_scale_object_query}) are generated based on the template image. Since no well-annotated training data is available for \revise{TIMOT}, we design a Dynamic Matching Training Strategy (DMTS, \Sref{sec_dynamic_matching_training_strategy}), which supports the training of Siamese-DETR on commonly used detection datasets (\emph{e.g.}, COCO~\cite{lin2014microsoft}). The provided annotations are fully utilized more than once when multiple template images are provided for training. During the inference stage, objects are tracked in a Track-by-Query (TbQ, \Sref{sec_tracking_by_query}) manner, as shown in \Fref{fig_motivation}. The tracked boxes in the previous frame are used as additional query boxes to track corresponding objects. A simple NMS operation, rather than the complex data association algorithm, is adopted to remove some duplicated boxes. To make Siamese-DETR compatible with such a tracking strategy, the query denoising~\cite{li2022dn} is adopted and optimized to train Siamese-DETR.

\subsection{Multi-scale Object Queries}
\label{sec_multi_scale_object_query}
Following previous works~\cite{liu2022dab,zhang2022dino,li2022dn,lai2024hyperspectral}, we use the decoupled object queries. Formally, let $Q = \{q_n| q_n = (\mathbf{q}_{c_n}, \mathbf{q}_{b_n}), n=0,1,...,N-1\}$ be the set of object queries, where $N$ is the number of queries. For each query $q_n = (\mathbf{q}_{c_n}, \mathbf{q}_{b_n})$, the query content $\mathbf{q}_{c_n} \in \mathbb{R}^{D}$ is a feature vector with dimensionality $D$, and the query box $\mathbf{q}_{b_n} \in \mathbb{R}^{4}$ is represented by the center coordinate, width and height. In DETR variants~\cite{zhu2020deformable,li2022dn, liu2022dab, zhang2022dino}, the query contents are usually a set of parameters that are learned by the model. Such design works for object detection with closed-set categories, but is not suitable for generic object detection/tracking, where the category of template image provided in the inference stage may be unseen in the training stage.

In order to detect all objects that share the same category with the template image, we get query contents from the template image. More specifically, using the features extracted from the template image as the query contents. Our hypothesis is that the query contents store the semantic information of objects, \emph{e.g.}, intra-category common ground, which is vital for object detection. On the other hand, objects in the same scene vary a lot in terms of scale even if they share the same category. To handle this, the multi-scale features are extracted from the template image and used as multi-scale query contents. \revise{Formally, let $F = \{\mathbf{f}_{s}|s=0,1,..., S-1\}$ be the set of multi-scale feature maps extracted by the backbone network (\emph{e.g.}, Swin Transformer~\cite{liu2021swin}) from the given template image. We first get the feature vectors $\hat{F} = \{\mathbf{\hat{f}}_{s}|s=0,1,..., S-1\}$ by spatially average pooling the feature maps with:
\begin{equation}
	\mathbf{\hat{f}}_{s} = {\rm AvgPool}(\mathbf{f}_{s}).
\end{equation}  For the $n$-th object query, its content $\mathbf{q}_{c_n}$ is determined by:
\begin{equation}
    \mathbf{q}_{c_n} = \mathbf{\hat{f}}_{n \ {\rm mod} \ S}, 
    \label{eq_template_feature_to_content_query}
\end{equation}
where $n \ {\rm mod} \ S$ is the index of the feature vectors in $\hat{F}$.} As for query boxes, they are a set of learnable parameters that are optimized in the training stage following previous works~\cite{liu2022dab,li2022dn}, which means that different template images share the same query boxes.

\subsection{Dynamic Matching Training Strategy}
\label{sec_dynamic_matching_training_strategy}
Different from traditional MOT, where well-annotated training data is provided, there is no available well-annotated training data for \revise{TIMOT} ~\cite{bai2021gmot}. The common practice is to train the model on external datasets, and then test it on evaluation benchmark (\emph{i.e.}, GMOT-40~\cite{bai2021gmot}). Existing method~\cite{bai2021gmot} uses multiple datasets to train the detector, including LaSOT~\cite{fan2021lasot}, GOT-10K~\cite{huang2019got} and COCO~\cite{lin2014microsoft}. However, the detection/tracking performance is far from satisfactory. In this paper, we design the Dynamic Matching Training Strategy (DMTS) for the training of Siamese-DETR on commonly used detection datasets. It will be seen in \Sref{sec_experiments} that Siamese-DETR surpasses existing method~\cite{bai2021gmot} by a large margin in terms of detection and tracking, even only been trained on COCO~\cite{lin2014microsoft}. The superiority of the dynamic matching training strategy comes from two aspects: 1) utilizing all annotations even if they belong to different categories; 2) utilizing all annotations more than once for each training step.

\subsubsection{Utilizing All Annotations}
\label{sec_utilize_all_annotations}
Let $A = \{a_k | a_k = (\mathbf{b}_k, c_k), k=0,1, ..., K-1 \}$ be the set of annotations for the input training image, where $\mathbf{b}_k \in \mathbb{R}^{4}$ and $c_k \in \mathbb{Z}$ are the bounding box and category ID of the $k$-th object. We randomly sample a category ID from $\{c_0, c_1,...,c_{K-1}\}$, which is used as the category of template image and denoted as $\hat{c}_t$. Given the category ID $\hat{c}_t$, the template image is cropped from another image in the training split. The corresponding annotations for the given template image and the category ID $\hat{c}_t$ are: 
\begin{equation}
    A^{\hat{c}_t} = \{a^{\hat{c}_t}_{k} | a^{\hat{c}_t}_{k}=(\mathbf{b}_k, \mathbbm{1}_{c_k,\hat{c}_t}), k=0,1,...,K-1\}, 
\end{equation}
where:
\begin{equation}
\mathbbm{1}_{c_k,\hat{c}_t} = 
    \begin{cases} 
        1 & \text{if } c_k = \hat{c}_t, \\ 
        0 & \text{else}. 
    \end{cases}
\end{equation}
As we can see, the boxes in $A^{\hat{c}_t}$ are divided into positive ($c_k=\hat{c}_{t}$) and negative ($c_k \neq \hat{c}_{t}$) samples, resulting a \emph{two-category} object detection task, as shown in \Fref{fig_training_stage}. However, there exists another naive setting that results in a \textit{single-category} object detection task: keeping the boxes that share the same category with the template image and removing the others. Under this setting, only the positive samples are utilized. 
Though the latter seems to be more intuitive than the former, it is not a good choice due to the fact that poorer detection performance is achieved since no negative samples are utilized for the training, which weakens the capability of the model to distinguish the interested objects from others.

\subsubsection{Utilizing All Annotations More Than Once}
\label{sec_utilize_all_annotations_more_than_once}

Considering the fact that the template image and the multi-scale-object queries take up a small proportion of the device memory when compared with the input image and the whole model, we can provide multiple template images during training. 

Let $\{\hat{c}_0,\hat{c}_1,...,\hat{c}_{T-1}\}$ be the randomly sampled category IDs for $T$ different template images. For each template image with category $\hat{c}_t$, the obtained multi-scale object queries are denoted as $Q^{\hat{c}_t}$, which is associated with the annotations $A^{\hat{c}_t}$. The $T$ groups \revise{of} object queries $\{Q^{\hat{c}_0}, Q^{\hat{c}_1},..., Q^{\hat{c}_{T-1}}\}$ \revise{contain a total number of $N \times T$ object queries, which are} concatenated and fed into transformer for object detection within once forward. Note that the interactions between different object queries are only allowed within each group, and the object queries in different groups cannot see each other. This can be simply implemented by providing an attention mask to self-attention layers in the transformer decoder.

\subsection{Tracking-by-Query}
\label{sec_tracking_by_query}

The common tracking-by-detection paradigm usually contains two stages: 1) Object detection. The interested objects are firstly detected by the detector; 2) Data association. The trajectories of objects are obtained by matching objects that come from different frames. However, performing data association properly is non-trivial since it involves the computation of affinity matrix between different objects, the setting of affinity threshold that prevents a wrong association, \emph{etc}. In this paper, we use the tracked boxes in the previous frame as additional query boxes to track the corresponding objects. The query denoising is optimized to mimic the tracking scenarios on static images.

\subsubsection{Tracked Boxes as Additional Query Boxes}
Let $B = \{\mathbf{\hat{b}}_m|m=0,1,..., M\}$ be the set of tracked boxes in the previous frame, we construct additional object queries as follows:
\revise{
\begin{equation}
\label{eq_tracking_query}
\begin{split}
\hat{Q} = \hat{Q}_0 \cup \hat{Q}_1\cup...\cup \hat{Q}_{S-1} ,
\end{split}
\end{equation}
where $\hat{Q}_s$ is the subset of additional object queries that are constructed for the $s$-th scale:
\begin{equation}
\label{eq_tracking_query_2}
\begin{split}
\hat{Q}_s = \{\hat{q}_{s,m}|\hat{q}_{s,m}=(\mathbf{\hat{f}}_s,\mathbf{\hat{b}}_m), m&=0,1,...,M-1 \}.
\end{split}
\end{equation}
}Different from object queries $Q$, which is responsible for object detection, $\hat{Q}$ is used for object tracking. \revise{The inspiration of this design is that the category-aware information is conveyed by the features from the template image and embedded into the query contents, and the tracked boxes in the previous frame are close enough to the corresponding objects in the current frame.} While tracking online, the object queries in $Q$ and $\hat{Q}$ are concatenated and fed into the transformer together. \revise{The object queries in $Q$ and $\hat{Q}$ detect and track objects simultaneously but independently. For object tracking, different object instances are distinguished by their corresponding tracked boxes. To avoid the interactions between the object queries in $Q$ and $\hat{Q}$, an attention mask is provided to each self-attention layer in the transformer decoder.}

For each box $\mathbf{\hat{b}}_m$, $S$ tracked boxes are obtained by object queries $\{\hat{q}_{0,m}, \hat{q}_{1,m},...,\hat{q}_{S-1,m}\}$. Among these $S$ tracked boxes,  the one that has the largest Intersection over Union (IoU) with $\mathbf{\hat{b}}_m$ is \revise{selected. If the classification score of the selected box is higher than the predefined confidence threshold, it will be} kept as the tracking result for $\mathbf{\hat{b}}_m$. \revise{Otherwise, the corresponding object is treated as a disappeared object.} Since object queries in $Q$ and $\hat{Q}$ detect and track objects independently, the detection boxes from object queries in $Q$ may be duplicated with that from object queries in $\hat{Q}$. Following the MOT method Tracktor~\cite{bergmann2019tracking}, the NMS operation is used to remove the duplicated detection boxes. \revise{The remained detection boxes that have higher classification scores than the predefined confidence threshold are treated as the newly appeared objects.}

\begin{figure}[t]
    \centering
    \includegraphics[width=\linewidth]{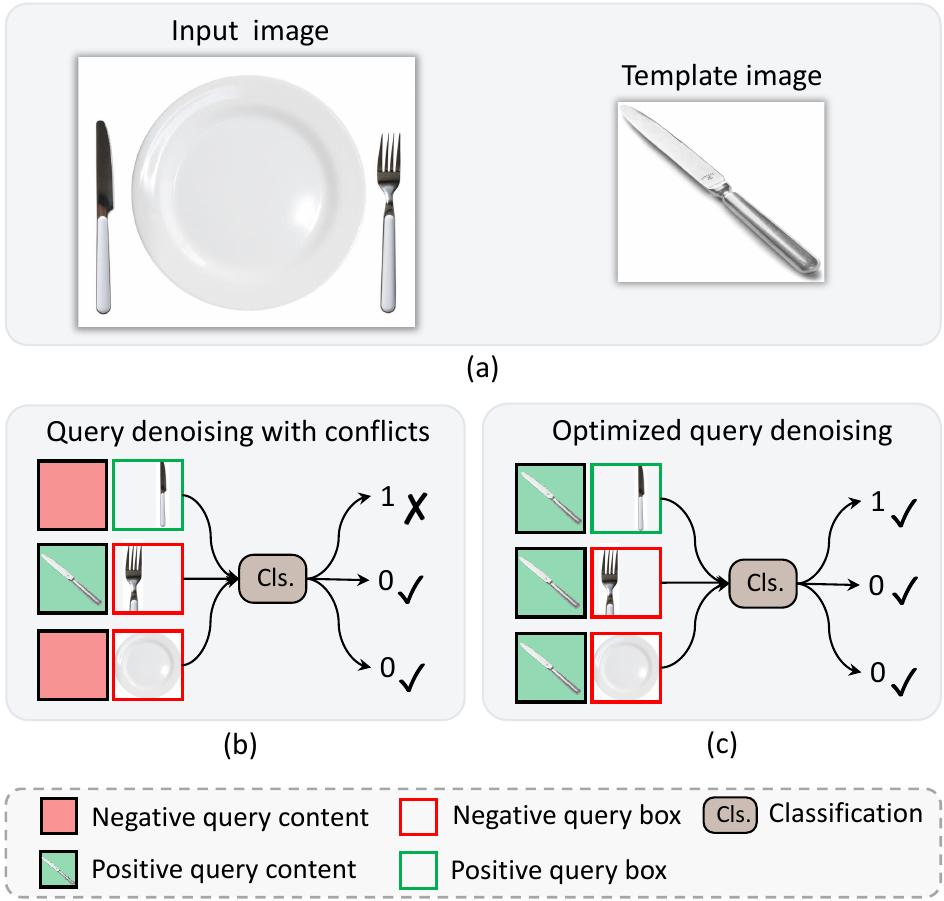}
    \caption{Illustration of query denoising. (a) Input image and template image. (b) Original query denoising~\cite{li2022dn} with conflicts for \revise{TIMOT}. The noisy object queries are classified according to the labeled category IDs that are associated with the query boxes, without taking the noisy query contents into consideration. (c) Optimized query denoising. The noisy object queries are classified according to the matching results between the query contents and noisy query boxes. \revise{The numbers 1 and 0 denote that the model tries to classify the object queries into positive and negative samples, while the markers \ding{55} and $\checkmark$ indicate whether the classification behaviors are wrong or right.}}
    \label{fig_query_denoising}
\end{figure}

\begin{figure*}[t]
\centering
\subfloat[\scriptsize Siamese-DETR + TbQ]{
\includegraphics[width=0.245\linewidth]{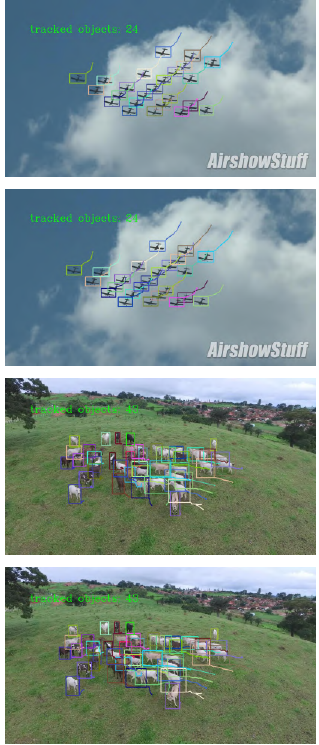}}
\subfloat[\scriptsize Siamese-DETR + SORT~\cite{bewley2016simple}]{
\includegraphics[width=0.241\linewidth]{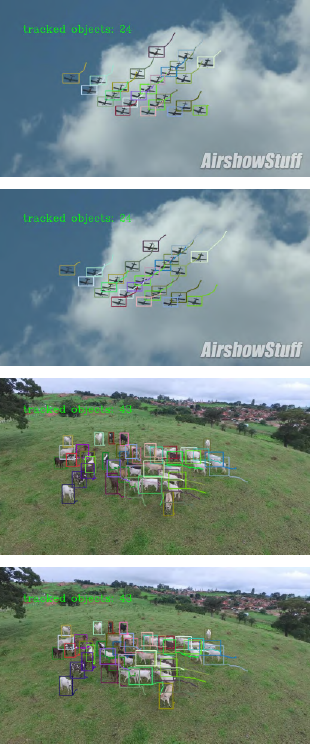}}
\subfloat[\scriptsize GLIP-T (B)~\cite{li2022grounded} + TbQ]{
\includegraphics[width=0.245\linewidth]{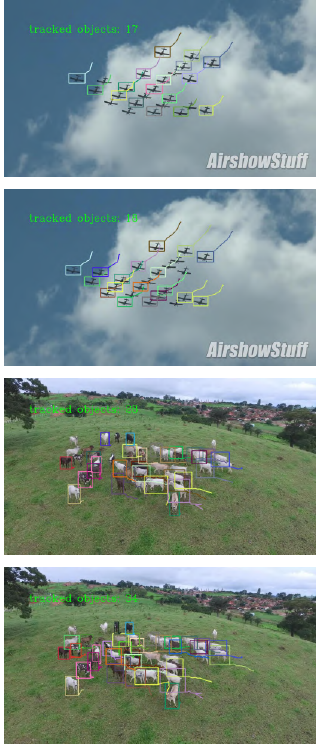}}
\subfloat[\scriptsize GLIP-T (B)~\cite{li2022grounded} + SORT~\cite{bewley2016simple}]{
\includegraphics[width=0.245\linewidth]{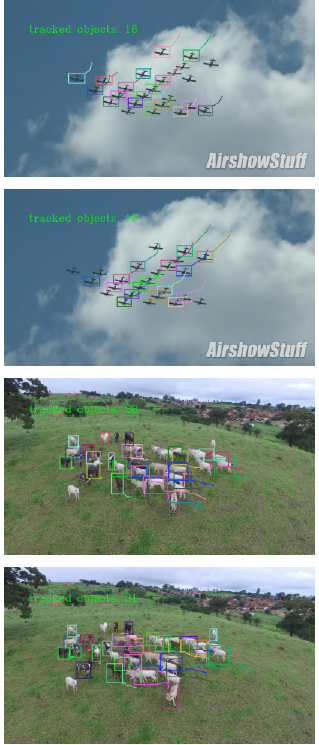}}
\caption{Qualitative comparison for different methods. The following two points can be summarized: 1) when combined with the same tracker (\textit{e.g.}, TbQ), our Siamese-DETR tracks more objects than GLIP-T (B)~\cite{li2022grounded}; 2) based on the same detector, our TbQ pipeline also tracks more objects than SORT~\cite{bewley2016simple}. The Siamese-DETR trained on COCO~\cite{lin2014microsoft} with \revise{Swin-T}~\cite{liu2021swin} as the backbone network is evaluated.
}
\label{fig_track_vis}
\end{figure*}

\subsubsection{Optimized Query Denoising}
\label{sec_optimized_query_denoising}
Since there is no well-annotated video data for training, it is hard for the model to track objects with tracked boxes while tracking online if the model is trained on static images. \revise{In addition, the number of interested objects varies from frame to frame while tracking online, which introduces more challenges for tracking.} To handle this, we add some noise to the ground-truth boxes, which are used as the tracked boxes (\textit{i.e.}, additional query boxes) during the training stage. The objects are \textit{tracked} by their corresponding noisy query boxes \revise{and Siamese-DETR is constrained to learn the capability of handling the various numbers of objects. With the mimicked tracking scenario in the training stage, Siamese-DETR is able to track objects with the proposed Tracking-by-Query strategy.} Similar to the online tracking stage, the groups of object queries for detection and tracking are independent of each other in the training stage.

The aforementioned strategy is similar to query denoising~\cite{li2022dn} which is commonly used in DETR variants. However, we find that existing query denoising is not suitable for \revise{TIMOT} task. The reason is that except for the box noise in query boxes, there exists category noise in query contents. Specifically, the category ID of a ground-truth box is randomly switched to another category ID. For each category ID, an embedding vector is learned by the detector and used as the noisy query content. While training, the noisy object queries are classified based on the labeled category IDs that are associated with the query box, without taking the query contents into consideration. There are two conflicts in existing query denoising when applied to Siamese-DETR: 1) Siamese-DETR performs \textit{two-category} detection task. Learning two embedding vectors for positive and negative samples is not suitable since the positive and negative samples are dynamically changed according to the provided template images (\textit{i.e.}, query contents); 2) A positive noisy query box may be paired with the negative query content, but still be classified as a positive sample in the original query denoising~\cite{li2022dn}. However, the object query is positive only when the query content and query box matched with each other in Siamese-DETR.

 To avoid these conflicts, we optimize query denoising by pairing all noisy query boxes with positive query contents (\emph{i.e.}, the features extracted from the template image). With this optimized query denoising, noisy object queries are classified correctly according to the matching results between query contents and query boxes. The difference between the original query denoising~\cite{li2022dn} and the optimized query denoising is illustrated in \Fref{fig_query_denoising}.

\subsection{\revise{Training of Siamese-DETR}}
\revise{The proposed Siamese-DETR is trained on the commonly used detection dataset, \textit{i.e.}, COCO~\cite{lin2014microsoft}. Given an image for training, we first randomly sample $T$ category IDs from the corresponding annotations. For each category ID, we then randomly sample a box from the annotations of another image, which is randomly sampled from the training split. Finally, the template images are cropped from the randomly sampled images with the sampled boxes.} \revise{For the $t$-th category ID $\hat{c}_t$,  except for the object queries $Q^{\hat{c}_t}$, we also construct additional object queries $\hat{Q}^{\hat{c}_t}=\hat{Q}^{\hat{c}_t}_0 \cup \hat{Q}^{\hat{c}_t}_1 \cup ... \cup \hat{Q}^{\hat{c}_{t}}_{S-1}$ with the noisy ground-truth boxes to mimic the tracking scenarios. The overall training loss is:
\begin{equation}
	\mathcal{L}=\frac{1}{T}\sum_{T=0}^{T-1}(\mathcal{L}_{\rm H}(A^{\hat{c_t}},\hat{A}^{Q^{\hat{c}_t}}) + \frac{1}{S} \sum_{s=0}^{S-1} \mathcal{L}_{\rm R}(A^{\hat{c}_t},\hat{A}^{\hat{Q}_{s}^{\hat{c}_t}}) )
\end{equation} 
where $\hat{A}^{Q^{\hat{c}_t}}$ and $\hat{A}^{Q^{\hat{c}_t}_s}$ denote the predictions from object queries in $Q^{\hat{c}_t}$ and $\hat{Q}_{s}^{\hat{c}_t}$, and $\mathcal{L}_{\rm H}(\cdot, \cdot)$ and $\mathcal{L}_{\rm R}(\cdot, \cdot)$ are the Hungarian loss~\cite{carion2020end} and reconstruction loss~\cite{li2022dn}, respectively. Note that while computing the loss, the ground-truth boxes without noise are used.}

\section{Experiments}
\label{sec_experiments}
In this section,  the involved datasets and metrics are firstly introduced, followed by the implementation details. Then, we compare Siamese-DETR with existing multi-object tracking methods. Finally, some discussions are provided to show the effectiveness of different components.

\subsection{Datasets and Metrics}
\revise{We follow the setup of template-image-based MOT task~\cite{bai2021gmot} to compare the proposed Siamese-DETR with other methods. Specifically, a tracker is tested on all videos in GMOT-40 benchmark~\cite{bai2021gmot} and can be trained on any other benchmark except GMOT-40.} \revise{It is worth noting that the template-image-based MOT trackers not only need to track all the objects of the same category with the template image in the video but also need to maintain the identity of each object.} \revise{For the reason that each video in GMOT-40 only contains one category, the Urban Tracker~\cite{jodoin2014urban} dataset is used to analyze the capability of Siamese-DETR in multi-category multi-object tracking.} In this work, the commonly used object detection dataset, COCO~\cite{lin2014microsoft}, is mainly used to train Siamese-DETR. Other datasets, for example, LVIS~\cite{gupta2019lvis} and Objects365~\cite{shao2019objects365}, are used for more detailed analysis. \revise{Though the categories in GMOT-40 and Urban Tracker may be visible in the training dataset (\textit{e.g.}, COCO, LVIS, Objects365), it complies with the setup of TIMOT~\cite{bai2021gmot}.}

\textbf{GMOT-40}  contains 40 videos that consist of 10 different object categories with 4 videos for each category. The entire dataset contains 9.6K frames, where 85.28\% of them contain more than 10 objects. The videos are shot with FPS ranging from 24 to 30. All the videos are used for evaluation.

\revise{\textbf{Urban Track} has 4 outdoor videos with each of them containing more than 1 category, resulting in a total number of 4 categories. The videos are captured with FPS of 25 or 30. All these 4 outdoor videos are used for evaluation.}

\textbf{COCO} is widely used for object detection. It contains a total number of 118K images and 860K annotated instances for training. There are 80 different categories in COCO, such as person, car, dog, and so on.  Note that only the category IDs and bounding boxes are used in this work.

\textbf{LVIS} shares nearly the same images with COCO but provides more fine-grained annotations. It contains 100K images and 1.27M instances for training. The number of annotated categories is 1203, providing more fine-grained category annotations than COCO. Note that the used bounding boxes are obtained from the annotated instance-level masks since there are no annotated bounding boxes in LVIS.

\textbf{Objects365} contains 0.6M images and 8.54M instances for training. The number of annotated categories is 365. We use Objects365 to show that more training data can boost the performance of Siamese-DETR.

We adopt the standard metrics of multi-object tracking for evaluation, including:
Multi-Object Tracking Accuracy (MOTA) \cite{bernardin2008evaluating},
IDentity F1 Score (IDF1),
Mostly Tracked objects (MT),
Mostly Lost objects (ML),
Number of False Positives (FP),
Number of False Negatives (FN) and
Number of Identity Switches (IDSw) \cite{li2009learning}.
Some other metrics, including mean Average Precision with IoU threshold 0.5 (mAP@0.5) and mean Average Recall (mAR) are also adopted for the evaluation of object detection.

\begin{table*}
\setlength{\tabcolsep}{2.5pt}
\caption{\revise{The details of compared detection methods. Note that the extra costs of FLOPs and inference time in the processing of the template image or text prompt are not included since they only need to be processed in the first frame and the costs are negligible when the number of frames is large enough. Time consumption is evaluated on a workstation with a 3.9GHz CPU and an RTX 3090 GPU. The best results are shown in bold with underline.}}
\label{table_details_of_detection_methods}
\centering	
\begin{tabular}{c|c|c|c|c|c|c|c|c} 
\hline
\makecell{\revise{Detection Methods}} &\revise{Backbone} &\revise{Neck}  &\makecell{\revise{Backbone Pre-}\\ \revise{training Dataset}} &\makecell{\revise{(Vision-)} \\ \revise{Lauguage} \\ \revise{Model}} &\makecell{\revise{Inference} \\ \revise{Resolution}} &\makecell{\revise{Parameter}\\\revise{(M)}} &\makecell{\revise{FLOPs} \\ \revise{(G)}} &\makecell{\revise{Inference} \\ \revise{Time (ms)}} \\
\hline 
\revise{YOLOv5l6~\cite{couturier2021deep}} & \revise{CSP-DarkNet53~\cite{redmon2018yolov3}} &\revise{SPPF~\cite{he2015spatial},CSP-PAN~\cite{liu2018path}} &\revise{$\times$} &\revise{$\times$} &\revise{1280$\times$1280} &\revise{76.8} &\revise{112.4} &\revise{\underline{\bf 16.7}} \\

\revise{DINO~\cite{zhang2022dino}} &\revise{ResNet50~\cite{he2016deep}} &\revise{DETR's Neck~\cite{carion2020end}}  &\revise{ImageNet~\cite{krizhevsky2012imagenet}} &\revise{$\times$}   &\revise{800$\times$1200} &\revise{45.2} &\revise{261.9} &\revise{87.5} \\

\revise{Conditional DETR~\cite{meng2021conditional}} &\revise{ResNet50~\cite{he2016deep}} &\revise{DETR's Neck~\cite{carion2020end}}&\revise{ImageNet~\cite{krizhevsky2012imagenet}}  &\revise{$\times$} &\revise{800$\times$1200} &\revise{43.2} &\revise{\underline{\bf 88.9}} &\revise{44.4}\\

\hline
\revise{OVTrack~\cite{li2023ovtrack}}  &\revise{ResNet50~\cite{he2016deep}} & \revise{Faster RCNN's Neck\cite{ren2015faster}}  &\revise{ImageNet~\cite{krizhevsky2012imagenet}} &\revise{CLIP~\cite{radford2021learning}} &\revise{800$\times$1333} &\revise{67.6} &\revise{191.2} &\revise{57.5} \\

\revise{GLIP-T (B)~\cite{li2022grounded}} &\revise{Swin-Tiny~\cite{liu2021swin}} &\revise{DyHead's Neck~\cite{dai2021dynamic}}&\revise{ImageNet~\cite{krizhevsky2012imagenet}} &\revise{BERT~\cite{devlin2018bert}} & \revise{800$\times$1333} &\revise{195.2} &\revise{322.4} &\revise{158.7} \\

\hline

\revise{GlobalTrack~\cite{huang2020globaltrack}} &\revise{ResNet50~\cite{he2016deep}} & \revise{Faster RCNN's Neck\cite{ren2015faster}} &\revise{ImageNet~\cite{krizhevsky2012imagenet}} &\revise{$\times$} &\revise{ 800$\times$1333} &\revise{\underline{\bf 41.3}} &\revise{169.9} &\revise{53.7 } \\

\revise{Siamese-DETR (Ours,Swin-T)} &\revise{Swin-Tiny~\cite{liu2021swin}} &\revise{DETR's Neck~\cite{carion2020end}} &\revise{ImageNet~\cite{krizhevsky2012imagenet}}  &\revise{$\times$} &\revise{800$\times$1200} &\revise{47.6} &\revise{267.3} &\revise{86.6} \\ 

\revise{Siamese-DETR (Ours,Swin-B)} &\revise{Swin-Base~\cite{liu2021swin}} &\revise{DETR's Neck~\cite{carion2020end}} &\revise{ImageNet~\cite{krizhevsky2012imagenet}}  &\revise{$\times$} &\revise{800$\times$1200} &\revise{108.2} &\revise{542.7} &\revise{139.9} \\

\hline
\end{tabular}
\end{table*}

\begin{table}[t]
\setlength{\tabcolsep}{2pt}
\caption{\revise{The details of evaluated tracking methods. The inference times of different methods are evaluated based on the detection results of Siamese-DETR (Swin-T). Time consumption is evaluated on a workstation with a 3.9GHz CPU and an RTX 3090 GPU, and the detection time consumption is excluded. The best results are shown in bold with underline.}}
\label{table_details_of_trackers}
\centering
\begin{tabular}{c|ccccc} 
\hline
\revise{Tracking Methods} & \revise{Online} &\makecell{\revise{Kalman} \\ \revise{Filter}} &\makecell{\revise{Appearance} \\ \revise{Cues}} &\makecell{\revise{Hierarchical} \\ \revise{Matching}} & \makecell{\revise{Inference} \\\revise{Time (ms)}} \\
\hline
\revise{IOU~\cite{bochinski2017high}} &\revise{$\times$} &\revise{$\times$} &\revise{$\times$} &\revise{$\times$} &\revise{$<$\underline{\bf 0.01}} \\
\revise{SORT~\cite{bewley2016simple}} &\revise{$\checkmark$} &\revise{$\checkmark$} &\revise{$\times$} &\revise{$\times$} &\revise{0.1} \\
\revise{DeepSORT~\cite{wojke2017simple}} &\revise{$\checkmark$} &\revise{$\checkmark$} &\revise{$\checkmark$} &\revise{$\checkmark$} &\revise{87.3}\\
\revise{ByteTrack~\cite{zhang2022bytetrack}} &\revise{$\checkmark$} &\revise{$\checkmark$} &\revise{$\checkmark$} &\revise{$\checkmark$} &\revise{33.7} \\
\revise{BoT-SORT~\cite{aharon2022bot}} &\revise{$\checkmark$} &\revise{$\checkmark$} &\revise{$\times$}  &\revise{$\checkmark$} &\revise{9.5}  \\
\revise{TbQ (Ours, Swin-T)} &\revise{$\checkmark$} &\revise{$\times$} &\revise{$\times$} &\revise{$\times$} &\revise{2.2} \\
\hline
\end{tabular}
\end{table}

\begin{table*}[!htbp]
\setlength{\tabcolsep}{2pt}
\belowrulesep=0pt
\aboverulesep=-1pt
\caption{Comparison with different methods on the GMOT-40 benchmark. Except for the \revise{TIMOT} methods, the methods with closed-set detectors and open-vocabulary detectors are also evaluated. Since different methods follow the tracking-by-detection pipeline, we divide them into the combination of a detector and a tracker. While combing our TbQ tracking pipeline with other detectors, the detection results provided by the evaluated detectors are fed into Siamese-DETR frame-by-frame, where the set of object queries $Q$ is removed and only $\hat{Q}$ is used for tracking. Both detection and tracking results are presented for a more comprehensive comparison. The best results are shown in bold with underline.}
\label{table_comparison_with_different_methods}
\centering
\begin{tabular}{c|c|c|c|cc|l|ccccccc} 
\hline
&\multirow{2}*{\makecell{Detection Methods}} & \multirow{2}*{\makecell{Language\\Models}} & \multirow{2}*{\makecell{Training Datasets for\\Detection Methods}} & \multicolumn{2}{c|}{Dectection Results} & \multicolumn{1}{c|}{\multirow{2}*{Tracking Methods}} &\multicolumn{7}{c}{Tracking Results} \\
\cline{5-6}\cline{8-14}
&& & & mAP@0.5$\uparrow$& mAR$\uparrow$ & & MOTA$\uparrow$& IDF1$\uparrow$& MT$\uparrow$& ML$\downarrow$& FP$\downarrow$& FN$\downarrow$& IDSw $\downarrow$ \\
\hline
\multirow{21}*{\rotatebox{90}{Closed-Set MOT Methods}} &&&&& &\multicolumn{8}{l}{\revise{YOLOv5l6} (manual)}\\

\cmidrule(lr){7-14}
&\multirow{6}{*}{\makecell{\revise{YOLOv5l6}~\cite{couturier2021deep}\\(manual)}} &\multirow{6}{*}{$\times$}  & \multirow{6}{*}{COCO~\cite{lin2014microsoft}} &\multirow{6}{*}{41.1\%} &\multirow{6}{*}{38.6\%} &  + IOU~\cite{bochinski2017high} &22.7\% &29.0\% &257 &1346 &38752 &161207 &4861\\
&& & & & & + SORT~\cite{bewley2016simple} &22.9\% &31.2\% &232 &1311 &39574 &162323 &2498 \\
&& & & & & + DeepSORT~\cite{wojke2017simple} &23.0\% &32.4\% &269 &1283 &40129 &159833 &2274 \\
&& & & & & + ByteTrack~\cite{zhang2022bytetrack} &23.6\% &34.4\% &301 &1261 &38763 &162037 &3691 \\
&& & & & & + BoT-SORT~\cite{aharon2022bot} &23.9\% &35.1\% &298 &1214 &37991 &158429 &3437\\
&& & & & & + TbQ (Ours, Swin-T) &24.5\% &25.1\% &284 &1241 &40032 &152860 &8431 \\
\cline{2-14}
&&&&&&\multicolumn{8}{l}{DINO (manual)}\\
\cmidrule(lr){7-14}
&\multirow{6}{*}{\makecell{DINO~\cite{zhang2022dino}\\(manual)}} &\multirow{6}{*}{$\times$} & \multirow{6}{*}{COCO~\cite{lin2014microsoft}} &\multirow{6}{*}{28.7\%} &\multirow{6}{*}{30.1\%} & + IOU~\cite{bochinski2017high} &21.7\% &29.0\% &257 &1346 &38752 &161207 &1789\\
&& & & & & + SORT~\cite{bewley2016simple} &19.3\% &23.1\% &182 &1437 &41786 &167923 &1658\\
&& & & & & + DeepSORT~\cite{wojke2017simple} &20.0\% &23.4\% &184 &1390 &42013 &165689 &1931\\
&& & & & & + ByteTrack~\cite{zhang2022bytetrack} &21.2\% &25.7\% &201 &1376 &43543 &162480 &1774\\
&& & & & & + BoT-SORT~\cite{aharon2022bot} &21.1\% &26.0\% &199 &1345 &42081 &163093 &3762\\
&& & & & & + TbQ (Ours, Swin-T) &21.9\% &24.1\% &215 &1247 &41678 &163881 &6597 \\

\cline{2-14}
&&&&&&\multicolumn{8}{l}{\revise{Conditional DETR (manual)}}\\
\cmidrule(lr){7-14}
&\multirow{6}{*}{\makecell{\revise{Conditional} \\\revise{DETR~\cite{meng2021conditional}}\\\revise{(manual)}}} &\multirow{6}{*}{\revise{$\times$}} & \multirow{6}{*}{\revise{COCO~\cite{lin2014microsoft}}} &\multirow{6}{*}{\revise{33.1\%}} &\multirow{6}{*}{\revise{22.0\%}} & \revise{+ IOU~\cite{bochinski2017high}}& \revise{17.7\%}& \revise{20.5\%}& \revise{161} & \revise{1316} & \revise{21109} & \revise{184826} & \revise{5056} \\
&& & & & & \revise{+ SORT~\cite{bewley2016simple}}& \revise{16.7\%} & \revise{22.7\%} & \revise{111} & \revise{1423} & \revise{15595} & \revise{195583} & \revise{2283} \\
&& & & & & \revise{+ DeepSORT~\cite{wojke2017simple}}& \revise{19.3\%} & \revise{28.7\%} & \revise{179} & \revise{1296} & \revise{23141} & \revise{180550} & \revise{3094}\\
&& & & & & \revise{+ ByteTrack~\cite{zhang2022bytetrack}}& \revise{17.9\%} & \revise{30.9\%} & \revise{183} & \revise{1129} & \revise{32922} & \revise{174627} & \revise{3964} \\
&& & & & & \revise{+ BoT-SORT~\cite{aharon2022bot}}& \revise{19.1\%} & \revise{34.2\%} & \revise{232} & \revise{1093} & \revise{32866} & \revise{170789} & \revise{3716} \\
&& & & & & \revise{+ TbQ (Ours, Swin-T)}& \revise{19.2\%} & \revise{24.7\%} & \revise{230} & \revise{1075} & \revise{31317} & \revise{163955} & \revise{9499}\\

\hline
\multirow{14}*{\rotatebox{90}{Open-Vocabulary MOT Methods}}&&&&&& \multicolumn{8}{l}{OVTrack}\\
\cmidrule(lr){7-14}
&\multirow{6}{*}{\makecell{OVTrack~\cite{li2023ovtrack}}} &\multirow{6}{*}{$\checkmark$} & \multirow{6}{*}{\makecell{LVIS~\cite{gupta2019lvis}}} &\multirow{6}{*}{31.7\%} &\multirow{6}{*}{32.7\%} & + IOU~\cite{bochinski2017high} &20.3\% &18.8\% &139 &1257 &50467 &154367 &1473\\
&& & & & & + SORT~\cite{bewley2016simple} &18.9\% &20.1\% &145 &1387 &49850 &158905 &1578\\
&& & & & & + DeepSORT~\cite{wojke2017simple} &20.2\% &21.2\% &165 &1367 &49984 &160378 &1470\\
&& & & & & + ByteTrack~\cite{zhang2022bytetrack} &19.9\% &20.6\% &164 &1345 &51356 &156329 &1669\\
&& & & & & + BoT-SORT~\cite{aharon2022bot} &20.0\% &20.3\% &167 &1328 &45721 &163378 &3278\\
&& & & & & + TbQ (Ours, Swin-T) &21.3\% &18.7\% &186 &1304 &50784 &154893 &6381 \\
\cline{2-14}
&&&&&& \multicolumn{8}{l}{GLIP-T (B)}\\
\cmidrule(lr){7-14}
&\multirow{6}{*}{\makecell{GLIP-T (B)~\cite{li2022grounded}}} &\multirow{6}{*}{$\checkmark$} & \multirow{6}{*}{\makecell{Objects365~\cite{shao2019objects365}}} &\multirow{6}{*}{50.8\%} & \multirow{6}{*}{44.6\%} & + IOU~\cite{bochinski2017high} &25.1\% &39.3\% &458 &721 &55802 &139560 &6320\\
&& & & & & + SORT~\cite{bewley2016simple} &25.2\% &40.8\% &354 &877 &54891 &143623 &2987\\
&& & & & & + DeepSORT~\cite{wojke2017simple} &25.5\% &41.6\% &401 &877 &46610 &141330 &2892 \\
&& & & & & + ByteTrack~\cite{zhang2022bytetrack} &27.0\% &45.1\% &447 &746 &52591 &131759 &2706 \\
&& & & & & + BoT-SORT~\cite{aharon2022bot} &27.3\% &49.1\% &553 &643 &51308 &133462 &4675\\ 
&& & & & & + TbQ (Ours, Swin-T) &27.5\% &39.8\% &581 &592 &49470 &136602 &9972  \\
\cline{2-14}

\hline
\multirow{16}*{\rotatebox{90}{\revise{Template-Image-based MOT Methods}}}&&&&&& \multicolumn{8}{l}{GlobalTrack}\\
\cmidrule(lr){7-14}
&\multirow{6}{*}{\makecell{GlobalTrack~\cite{huang2020globaltrack}}} &\multirow{6}{*}{$\times$} & \multirow{6}{*}{\makecell{COCO~\cite{lin2014microsoft}, \\LaSOT~\cite{fan2021lasot}, \\GOT-10K~\cite{huang2019got}}} &\multirow{6}{*}{28.3\%} & \multirow{6}{*}{18.3\%} & + IOU~\cite{bochinski2017high} & 11.8\%& 20.3\%& 56& 1491& \underline{\bf 8299} & 216821& 1668\\
&& & & & & + SORT~\cite{bewley2016simple} & 19.5\%& 30.3\%& 140& 1187& 15132& 189315& 1785 \\
&& & & & & + DeepSORT~\cite{wojke2017simple} & 14.5\%& 24.4\% &72 & 1363& 9000& 208818& \underline{\bf 1315} \\
&& & & & & + ByteTrack~\cite{zhang2022bytetrack} &19.1\% &32.1\% &178 &1069 &23881 &181829 &1791 \\
&& & & & & + BoT-SORT~\cite{aharon2022bot} &19.4\% &34.0\% &251 &978 &22229 &176991 &7375 \\
&& & & & & + TbQ (Ours, Swin-T) &20.6\% &27.4\% &213 &1066 &13507 &182376 &6407  \\
\cline{2-14}
&&&&&& \multicolumn{8}{l}{Siamese-DETR (Ours, Swin-T)}\\
\cmidrule(lr){7-14}
&\multirow{6}{*}{\makecell{Siamese-DETR\\(Ours, Swin-T)}} &\multirow{6}{*}{$\times$}  & \multirow{6}{*}{\makecell{COCO~\cite{lin2014microsoft}}} &\multirow{6}{*}{57.5\%} & \multirow{6}{*}{46.6\%} & + IOU~\cite{bochinski2017high} & 30.7\%& 35.0\%& 361& 759& 42504& 127024& 8158\\
&& & & & & + SORT~\cite{bewley2016simple} &30.1\%& 34.7\%& 235& 943& 29518& 145613& 4060 \\
&& & & & & + DeepSORT~\cite{wojke2017simple} &31.1\%& 41.8\%& 382& 773& 47336& 124257& 5131 \\
&& & & & & + ByteTrack~\cite{zhang2022bytetrack} &33.7\% & 41.4\%& 331& 764& 53417&104765 &4204  \\
&& & & & & + BoT-SORT~\cite{aharon2022bot} &34.1\%& 47.5\%& 431& 674& 45769& 119288& 6775 \\
&& & & & & + TbQ (Ours, Swin-T) &35.9\% &42.8\% &504 &666 &44882 &107894 &11664\\

\cline{2-14}
&\multirow{2}{*}{\makecell{Siamese-DETR\\(Ours, Swin-B)}} &\multirow{2}{*}{$\times$} &\multirow{1}{*}{COCO~\cite{lin2014microsoft}} &\multirow{1}{*}{63.3\%} &\multirow{1}{*}{49.9\%} & + TbQ (Ours, Swin-B) &39.4\% &35.7\% &482 &586 &33968 &106079 &10233  \\
\cline{4-14}
&&&\multirow{1}{*}{Objects365~\cite{shao2019objects365}} &\multirow{1}{*}{\underline{\bf 69.6\%}} &\multirow{1}{*}{\underline{\bf 55.4\%}} &+  TbQ (Ours, Swin-B) &\underline{\bf 50.0\%} &\underline{\bf 51.3\%} &\underline{\bf 1083} &\underline{\bf 278} &44390 &\underline{\bf 68189} &11252 \\

\hline
\end{tabular}
\end{table*}

\subsection{Implementation Details}
We use Swin Transformer~\cite{liu2021swin} as the backbone network. Like most DETR variants~\cite{li2022dn, zhang2022dino}, there are 6 encoder layers and 6 decoder layers in the transformer, in which the hidden dimensionality is set to 256. Following the settings in Deformable-DETR~\cite{zhu2020deformable}, the number of feature scales $S$ is set to 4. The number of object queries $N$ is set to 600. Without specification, all evaluated Siamese-DETR variants are optimized with AdamW\revise{~\cite{loshchilov2018decoupled}} for 12 epochs. The batch size is set to 16 and the number of templates $T$ is set to 7 by default. The initial learning rate is set to $5e^{10^{-5}}$, which is decayed by a factor of $0.1$ at epoch 11. The longer side of the template image is resized to 400 before being fed into the backbone network. 
\revise{While tracking online, the NMS threshold is set to 0.5 to remove the duplicated detection boxes and we follow the settings of TIMOT~\cite{bai2021gmot} to crop the template image from the first frame with a randomly sampled box for each video.}
In the following, Siamese-DETR denotes the model for object detection and tracking if there is no ambiguity, otherwise, we use Siamese-DETR and TbQ to denote the detector and tracker, respectively.

\subsection{Comparison with Existing Methods}
\label{sec_compare_with_existing_methods}

We compare the proposed method with several existing \revise{closed-set MOT methods, open-vocabulary MOT methods and template-image-based MOT methods. It is worth noting that the model weights of existing methods provided by the authors are directly used for evaluation, and it is practical for the reason that GMOT-40 benchmark is just used for evaluation and not used in the training of all methods (including ours). Since different types of MOT methods follow a tracking-by-detection pipeline, we divide each of the evaluated methods into a detector and a tracker. The details of different detectors and tracker are shown in \Tref{table_details_of_detection_methods} and \Tref{table_details_of_trackers}, respectively.} For a comprehensive comparison, we not only apply our TbQ tracking strategy to different detectors but also apply different trackers to our detector. The detection and tracking results of different methods on GMOT-40~\cite{bai2021gmot} are provided in  ~\Tref{table_comparison_with_different_methods}

\subsubsection{\revise{Comparison of Details}}
\revise{From \Tref{table_details_of_detection_methods}, we can see that different detectors are not strictly constrained to have the same backbone network, detection neck, \textit{etc}. Compared with closed-set detectors (\textit{i.e.}, YOLOv5l6~\cite{couturier2021deep}, DINO~\cite{zhang2022dino}, Conditional DETR~\cite{meng2021conditional}), the open-vocabulary detectors (\textit{i.e.}, OVTrack~\cite{li2023ovtrack} and GLIP-T~\cite{li2022grounded}) and template-image-based detectors (\textit{i.e.}, GlobalTrack~\cite{huang2020globaltrack} and Siamese-DETR) need extra costs of FLOPs and inference time to process the text prompts or template images. However, the extra costs are negligible when the number of frames is large enough since the text prompt and template image only need to be processed in the first frame. Note that the (vision-)language models are also necessary for open-vocabulary methods. Our Siamese-DETR (Swin-T) has almost the same number of parameters, FLOPs, and inference time as DINO. From \Tref{table_details_of_trackers}, it can be observed that the commonly used Kalman Filter, appearance cues and hierarchical matching strategies are not used in our TbQ. The time consumption of TbQ for tracking is much less than that of detection for the reason that TbQ only introduces a few additional object queries for tracking, which is performed simultaneously with detection.}

\subsubsection{Comparison of Detection Performance}
\label{sec_comparison_of_Detection_performance}
We first show the detection performance of different methods on GMOT-40~\cite{bai2021gmot}. For closed-set methods, \textit{e.g.}, YOLOv5~\cite{couturier2021deep} \revise{(specifically, YOLOv5l6)}, DINO~\cite{zhang2022dino} \revise{and Conditional DETR~\cite{meng2021conditional}}, they tend to detect all objects that belong to the categories in the pre-defined closed-set, which results in a poor detection performance due to the fact that only the objects of one specific category are treated as foreground. To make these closed-set methods compatible with the setting of \revise{TIMOT}, we manually remove the predicted boxes that do not have the same category with the template image for each video (denoted as \revise{YOLOv5l6} (manual), DINO (manual) \revise{and Conditional DETR (manual)}). As expected, Siamese-DETR (Swin-T) outperforms \revise{YOLOv5l6} (manual), DINO (manual) \revise{and Conditional DETR (manual)} by a large margin. For example, when trained on the same dataset (\textit{i.e.}, COCO\cite{lin2014microsoft}), Siamese-DETR (Swin-T) achieves 16.4\%, 28.8\% and \revise{24.4\%} higher mAP@0.5 than \revise{YOLOv5l6} (manual), DINO (manual) \revise{and Conditional DETR (manual)}, respectively. \revise{YOLOv5l6} (manual) performs better than DINO (manual) \revise{and Conditional DETR (manual)}. The reason is that the objects in GMOT-40 are much smaller than those objects in COCO \revise{and DETR-based detectors (DINO and Conditional DETR) cannot handle small objects well~\cite{carion2020end}. Interestingly, compared with DINO (manual), Conditional DETR (manual) achieves better mAP@0.5 but much poorer mAR, resulting in poorer tracking results.}

As for open-vocabulary methods, they can detect interested objects by providing different text prompts. However, due to the domain gap between vision and language, they need the well pre-trained language models to extract features from the given text prompts. In addition, in order to recognize the accurate objects from different text prompts, fine-grained category annotations are required. For example, GLIP-T (B)~\cite{li2022grounded} utilizes the pretrained BERT~\cite{devlin2018bert} to extract text features and the detection model is trained on Objects365~\cite{shao2019objects365}. The number of categories in Objects365 is increased from 365 to 1300+ by the authors to provide more fine-grained annotations. 
\revise{During testing, we follow the settings of OVTrack and GLIP-T (B) to use the category names as the text prompts for object detection.}
It can be observed that without the help of well pre-trained language model and the fine-grained category annotations, our Siamese-DETR (Swin-T) outperforms OVTrack and GLIP-T (B) by a large margin when only the COCO~\cite{lin2014microsoft} dataset is used for training.

The template-image-based method GlobalTrack~\cite{huang2020globaltrack} is originally designed for Single Object Tracking (SOT) task. \revise{It is implemented based on a traditional detector~\cite{ren2015faster}, and it can detect the interested objects within the whole image for the reason that it computes the correlation score between the features extracted from the template image and the whole image. A high score indicates a potential object. Though there are lots of SOT methods (\textit{e.g.}, transformer-based TransT~\cite{chen2021transformer} and TrDiMP~\cite{wang2021transformer}), they are not suitable for template-image-based MOT task for the reason that only a small search region image rather than the whole image is supported to track the single object. Directly feeding the whole image to these SOT methods produces poor detection results (\textit{i.e.}, mAP@0.5=0.0\%).} Though multiple datasets are used in GlobalTrack for training, poor performance is achieved. For example, our COCO-trained Siamese-DETR (Swin-T) achieves 29.2\% higher mAP@0.5 than GlobalTrack.

Lastly, we utilize a larger-scale backbone network and train Siamese-DETR with more data to show the scalability of Siamese-DETR. It can be seen that: 1) with the same training data (\textit{i.e.}, COCO), the mAP@0.5 is increased from 57.5\% to 63.3\% when the backbone network is switched from Swin-T to Swin-B; 2) With the same model (Siamese-DETR (Swin-B)), the mAP@0.5 is further improved to 69.3\% when Object365 is used for training. The results demonstrate that the detection performance of Siamese-DETR can be boosted by a larger scale of model or more training data.

\begin{table*}[!htbp]
\setlength{\tabcolsep}{2.5pt}
\caption{Impact of the number of scales in \revise{Multi-Scale} Object Queries (MSOQ). Except for the different number of scales in object queries, all models are trained with 1 template image (refer to \Sref{sec_utilize_all_annotations_more_than_once}) and the negative samples are removed (refer to \Sref{sec_utilize_all_annotations}). The query denoising (refer to \Sref{sec_optimized_query_denoising}) is not utilized. The best results are shown in bold with underline.}
\label{table_results_of_different_number_of_scales}
\centering
\begin{tabular}{c|cc|cc|cc|cc|ccccccc}
\hline
\multirow{3}{*}{\makecell{Number \\ of Scales}} & \multicolumn{8}{c|}{Detection Results} & \multicolumn{7}{c}{Tracking Results} \\
\cline{2-16}
& \multicolumn{2}{c|}{Small} & \multicolumn{2}{c|}{Medium} & \multicolumn{2}{c|}{Large} & \multicolumn{2}{c|}{Overall} &\multirow{2}*{MOTA$\uparrow$} & \multirow{2}*{IDF1$\uparrow$} & \multirow{2}*{MT$\uparrow$} & \multirow{2}*{ML$\downarrow$} & \multirow{2}*{FP$\downarrow$} & \multirow{2}*{FN$\downarrow$} & \multirow{2}*{IDSw $\downarrow$}\\
\cline{2-9}
& mAP@0.5$\uparrow$& mAR$\uparrow$ & mAP@0.5$\uparrow$& mAR$\uparrow$ & mAP@0.5$\uparrow$& mAR$\uparrow$ & mAP@0.5$\uparrow$& mAR$\uparrow$ \\
\hline
1 &11.0\% &23.4\% &37.7\% &49.8\% &42.3\% &58.7\% &35.9\% &28.4\% &19.3\% &18.5\% &132 &1097 &\underline{\bf 23213} &175968 &12622\\
2 &13.1\% &35.2\% &40.0\% &55.2\% &48.9\% &71.0\% &39.7\% &33.9\% &20.2\% &19.5\% &138 &1084 &24372 &170329 &12566\\
3 &16.9\% &36.8\% &42.9\% &57.5\% &50.2\% &71.2\% &41.6\% &35.5\% &20.5\% &20.4\% &146 &1069 &25761 &165839 &\underline{\bf 11893}\\ 
4 &\underline{\bf 23.3\%} &\underline{\bf 45.8\%} &\underline{\bf 45.9\%} &\underline{\bf 60.3\%} &\underline{\bf 55.4\%} &\underline{\bf 77.8\%} &\underline{\bf 47.9\%} &\underline{\bf 43.5\%} &\underline{\bf 23.4\%} &\underline{\bf 22.3\%} &\underline{\bf 157} &\underline{\bf 1001} &34583 &\underline{\bf 155681} &18562\\
\hline 
\end{tabular}
\end{table*}

\subsubsection{Comparison of Tracking Performance}
\label{sec_comparision_of_tracking_performance}
Firstly, we show the generalization of the proposed tracking strategy TbQ by applying it to different detectors. Specifically, while tracking online, the detection results of different detectors are fed into Siamese-DETR frame-by-frame, where the set of object queries $Q$ is removed and only $\hat{Q}$ is used for tracking. Since the tracking pipeline follows a tracking-by-detection paradigm, different tracking performances are achieved based on different detectors. Compared with GlobalTrack~\cite{huang2020globaltrack}, the MOTA is increased from 20.6\% to 35.9\% by our Siamese-DETR (Swin-T). However, we find that such promotion of Siamese-DETR (Swin-T) is mainly introduced by the lower FN metric. Based on this finding, we have tried to reduce the number of false negatives (FN) of GlobalTrack, but failed with the fact that GlobalTrack produces very low and similar confidence scores for most of the predicted boxes.

Then we apply different trackers to a specific detector. \revise{Taking} the proposed Siamese-DETR (Swin-T) for example, our TbQ achieves the best MOTA among all different trackers, even TbQ is much simpler than others \revise{(refer to \Tref{table_details_of_trackers})}. For example, TbQ achieves 35.9\% MOTA, which is higher than 33.7\% and 34.1\% MOTA scores that are achieved by ByteTrack~\cite{zhang2022bytetrack} and BoT-SORT~\cite{aharon2022bot}. It is worth noting that both ByteTrack and BoT-SORT are recently proposed trackers for traditional multi-object tracking and achieve remarkable tracking performance on MOTChallenge datasets\footnote{https://motchallenge.net/}. But they are very complex and contain lots of hyper-parameters. For example, the confidence scores and matching thresholds for the two-stage matching strategy. All these hyper-parameters are well-tuned for pedestrian tracking and they are even tuned for each video. During our experiments, directly using the parameters tuned for MOTChallenge on GMOT-40 produces very poor tracking performance (\textit{i.e.}, MOTA $<$ 0). Though we have tried our best to tune these parameters for GMOT-40, the tracking performances of ByteTrack and BoT-SORT still lag behind that of TbQ. This may be caused by the domain gap between GMOT-40 and MOTChallenge datasets. Different from ByteTrack and BoT-SORT, our TbQ tracks objects without bells and whistles but achieves better overall tracking performance (specifically, MOTA). 
 
 \revise{Through deeper analysis, we find that TbQ sometimes produces worse IDSw or IDF1 scores than other tracking methods. For example, while applying TbQ and SORT to GlobalTrack,  the IDSw score of TbQ is higher than that of SORT (6407 vs. 1315) and the IDF1 score of TbQ is lower than that of SORT (27.4\% vs. 30.3\%). The reasons are twofold: 1) IDSw and IDF1 are related to the number of tracked objects and trajectory segments. Since TbQ tracks more objects (higher MT), it potentially produces a higher IDSw score and a lower IDF1 score;} \revise{2) TbQ has inferior discriminability than existing tracking methods. However, this is reasonable for the reason that TbQ tracks objects without bells and whistles.  For example, the common practices of Kalman Filter, appearance cues and hierarchical matching are not used.}

Some qualitative tracking results are shown in \Fref{fig_track_vis}. It can be seen that Siamese-DETR (Swin-T) tracks more interested objects than GLIP-T (B) when combined with the same tracker and TbQ tracks more objects than SORT when the same detection results are used, demonstrating the effectiveness of our Siamese-DETR and TbQ.

\subsection{Dicussions}
In the following, Swin-T~\cite{liu2021swin} is used as the backbone network without specification.

\subsubsection{Multi-Scale Object Queries}
In Siamese-DETR, the multi-scale features extracted from the template image are used as the query contents in order to detect different scales of objects that share the same category with the template image. To show the effectiveness of MSOQ, we design different counterparts that have different numbers of scales, \textit{i.e.}, $S\in \{1, 2, 3, 4\}$. For a specific $S$, \revise{the $S$ feature maps that have the smallest spatial size are used}. The results are shown in \Tref{table_results_of_different_number_of_scales}. As we can see,  both detection and tracking performances are improved when more scales of features are used. Specifically, compared with the results of 1-scale object queries, 4-scale object queries boost the mAP@0.5 and MOTA by 12.0\% and 4.1\%. With the help of multi-scale features, objects of different scales are more easily to be detected and recognized. For example, when the number of scales is increased from 1 to 4, the mAP@0.5 scores for small, medium and large objects are improved by 12.3\%, 8.2\% and 13.1\%, respectively.

\begin{table*}[!htbp]
\setlength{\tabcolsep}{10.2pt}
\caption{Impact of Dynamic Matching Training Strategy (DMTS). The models are trained without query denoising (refer to \Sref{sec_optimized_query_denoising}). The best results are shown in bold with underline.}
\label{table_results_of_different_training_strategy_swin}
\centering
\begin{tabular}{cc|cc|ccccccc}
\hline
 \multicolumn{2}{c|}{DMTS} &\multicolumn{2}{c|}{Detection Results} & \multicolumn{7}{c}{Tracking Results} \\
\cline{1-11}
\makecell{Utilizing all \\ annotations} &\makecell{Number of \\ templates} & mAP@0.5$\uparrow$& mAR$\uparrow$& MOTA$\uparrow$& IDF1$\uparrow$& MT$\uparrow$ & ML$\downarrow$ & FP$\downarrow$& FN$\downarrow$ &IDSw$\downarrow$ \\
\hline
$\times$&1 &47.9\% &43.5\% &23.4\% &22.3\% &157 &1001 &34583 &155681 &18562\\
\hline
 \checkmark&1 &48.6\% &43.1\% &23.7\% &21.3\% &142 &1031 &33583 &156411 &18055\\
 \checkmark&2 &46.0\% &41.3\% &22.7\% &21.1\% &137 &1072 &33861 &158935 &18239\\
 \checkmark&3 &50.9\% &44.4\% &24.6\% &22.4\% &186 &953 &32402 &150154 &17715\\
 \checkmark&4 &52.5\% &44.9\% &26.3\% &24.3\% &283 &872 &\underline{\bf 30374} &142489 &16993\\
 \checkmark&5 &53.2\% &45.1\% &26.6\% &24.4\% &299 &\underline{\bf 831} &32712 &139210 &16918\\
 \checkmark&6 &52.7\% &43.9\% &26.3\% &24.3\% &284 &875 &30384 &141489 &16974\\
 \checkmark&7 &\underline{\bf 54.9\%} &\underline{\bf 46.3\%} &\underline{\bf 27.8\%} &\underline{\bf 25.1\%} &\underline{\bf 316} &764 &33169 &\underline{\bf 133245} &17542\\
 \checkmark&8 &51.2\% &44.3\% &24.6\% &23.4\% &286 &862 &35027 &142154 &17366\\
 \checkmark&9 &53.1\% &44.5\% &26.2\% &24.8\% &308 &842 &31600 &140580 &\underline{\bf 16695}\\
\hline
\end{tabular}
\end{table*}

\begin{table*}[!htbp]
\setlength{\tabcolsep}{7.2pt}
\caption{Effectiveness of query denoising. The models are trained with DMTS, where the number of template images is set to 7. The best results are shown in bold with underline.}
\label{table_results_of_query_denoising}
\centering
\begin{tabular}{c|cc|ccccccc|cc}
\hline
\multirow{2}*{\makecell{Query \\ denoising}} &\multicolumn{2}{c|}{Detection Results} & \multicolumn{7}{c}{Tracking Results} & \multicolumn{2}{|c}{GT boxes as Query Boxes}\\
\cline{2-12}
& mAP@0.5$\uparrow$& mAR$\uparrow$& MOTA$\uparrow$& IDF1$\uparrow$ & MT$\uparrow$ & ML$\uparrow$ & FP$\downarrow$& FN$\downarrow$ &IDSw$\downarrow$ & Avg. Conf.$\uparrow$ & Avg. IoU$\uparrow$ \\
\hline

  $\times$  &54.9\% &46.3\% &27.8\% &25.1\% &316 &764 &33169 &133245 &17542 &0.101 &\revise{0.021} \\
 original~\cite{zhang2022dino} &55.4\% &46.3\% &28.4\% &27.1\% &319 &\underline{\bf 837} &\underline{\bf 32847} &137591 &13197 &0.093 &\revise{0.103}\\    
 Optimized (Ours) &\underline{\bf 57.5\%} &\underline{\bf 46.6\%} &\underline{\bf 35.9\%} &\underline{\bf 42.8\%} &\underline{\bf 504} &666 &44882 &\underline{\bf 107894} &\underline{\bf 11664} &\underline{\bf 0.426} &\revise{\underline{\bf 0.730}}\\

\hline
\end{tabular}
\end{table*}

\begin{table*}[!htbp]
\setlength{\tabcolsep}{8pt}
\caption{Impact of different training datasets. The models are equipped with MSOQ (\Sref{sec_multi_scale_object_query}) and trained with DMTS (7 template images, \Sref{sec_dynamic_matching_training_strategy}) and optimized query denoising (\Sref{sec_optimized_query_denoising}). The best results are shown in bold with underline.}
\label{table_results_of_different_training_datasets}
\centering
\begin{tabular}{c|c|cc|ccccccc} 
\hline
 \multirow{2}*{Methods} & \multirow{2}*{Datasets} &\multicolumn{2}{c|}{Detection Results} & \multicolumn{7}{c}{Tracking Results} \\
\cline{3-11}
& & mAP@0.5$\uparrow$& mAR$\uparrow$& MOTA$\uparrow$& IDF1$\uparrow$& MT$\uparrow$ & ML$\downarrow$ & FP$\downarrow$& FN$\downarrow$ &IDSw$\downarrow$ \\
\hline
 \multirow{3}{*}{\makecell{GLIP-T (B)~\cite{li2022grounded}\\ + TbQ (Ours, Swin-T)}} & COCO~\cite{lin2014microsoft} &30.5\% &23.4\% &19.6\% &18.4\% &108 &1436 &\underline{\bf 24067} &189542 &\underline{\bf 3671}\\
 & LVIS~\cite{gupta2019lvis} &38.8\% &30.2\% &22.3\% &21.3\% &116 &1361 &26478 &178435 &6549\\
 & Objects365~\cite{shao2019objects365} & 50.8\% & 44.6\% &27.5\% &39.8\% &581 &592 &49470 &136602 &9972\\
 \hline
\multirow{3}*{\makecell{Siamese-DETR \\ (Ours, Swin-T)}} & COCO~\cite{lin2014microsoft} &57.5\% &46.6\% &35.9\% &42.8\% &504 &666 &44882 &107894 &11664\\
 & LVIS~\cite{gupta2019lvis} &56.5\% &42.5\% &30.3\% &39.5\% &845 &303 &32471 &136960 &14653\\
 & Objects365~\cite{shao2019objects365} &59.3\% &48.0\% &40.8\% &44.2\% &668 &519 &39300 &98199 &14186\\
 \hline
\multirow{3}*{\makecell{Siamese-DETR \\ (Ours, Swin-B)}} & COCO~\cite{lin2014microsoft} & 65.6\% &50.6\% &43.1\% &46.6\% &681 &466 &49478 &84765 &13723\\
 & LVIS~\cite{gupta2019lvis} &62.4\% &45.0\% &33.8\% &41.1\% &415 &562 &38285 &119800 &12312\\
 & Objects365~\cite{shao2019objects365} &\underline{\bf 69.6\%} &\underline{\bf 55.4\%} &\underline{\bf 50.0\%} &\underline{\bf 51.3\%} &\underline{\bf 1083} &\underline{\bf 278} &44390 &\underline{\bf 68189} &11252\\
 \hline
\end{tabular}
\end{table*}

\begin{table*}[!htbp]
\setlength{\tabcolsep}{7.5pt}
\caption{\revise{Tracking results of different methods. Due to the fact that DETR-based TrackFormer is a closed-set tracking method and is mainly designed for pedestrian tracking, only the pedestrian/person videos in GMOT-40~\cite{bai2021gmot} are used for evaluation. The best results are shown in bold with underline.}}
\label{table_compare_with_detr_based_mot_methods}
\centering
\begin{tabular}{c|c|cc|c|cccccc} 
\hline
 \multirow{2}*{\revise{Methods}} & \multirow{2}*{\makecell{\revise{Datasets for} \\ \revise{Training}}} &\multicolumn{2}{c|}{\revise{Detection Results}} & \multicolumn{7}{c}{\revise{Tracking Results}} \\
\cline{3-11}
& & \revise{mAP@0.5$\uparrow$}& \revise{mAR$\uparrow$}& \revise{MOTA$\uparrow$}& \revise{IDF1$\uparrow$}& \revise{MT$\uparrow$} & \revise{ML$\downarrow$} & \revise{FP$\downarrow$} & \revise{FN$\downarrow$} &\revise{IDSw$\downarrow$} \\
\hline
\revise{TrackFormer~\cite{meinhardt2022trackformer}} &\revise{MOT17~\cite{milan2016mot16}}& \revise{28.4\%} & \revise{31.8\%} &  \revise{-8.5\%} & \revise{\underline{\bf 36.4\%}} & \revise{39} & \revise{45} & \revise{13500} & \revise{10990} & \revise{\underline{\bf 86}}\\
\hline
\multirow{1}*{\revise{Siamese-DETR (Ours, Swin-T)}} &\multirow{1}*{\revise{COCO~\cite{lin2014microsoft}}}& \multirow{1}*{\revise{\underline{\bf 73.1\%}}} & \multirow{1}*{\revise{\underline{\bf 62.3\%}}} & \revise{\underline{\bf 43.7\%}} & \revise{29.5\%} & \revise{\underline{\bf 43}} & \revise{\underline{\bf 12}} & \revise{\underline{\bf 4661}} & \revise{\underline{\bf 6493}} & \revise{1609} \\
\hline
\end{tabular}
\end{table*}

\subsubsection{Dynamic Matching Training Strategy}
The dynamic matching training strategy is designed to efficiently train Siamese-DETR on commonly used detection datasets through \emph{utilizing all annotations} and \emph{utilizing all annotations more than once}. The results are shown in \Tref{table_results_of_different_training_strategy_swin}. When all annotations are used, Siamese-DETR performs a \textit{two-category} object detection task. The mAP@0.5 and MOTA scores are improved by 0.7\% and 0.3\%, respectively. However, mAR is reduced by 0.4\%, which is reasonable since an object is more potentially to be classified as the background when the negative samples are introduced to train the model (refer to \Sref{sec_dynamic_matching_training_strategy}).

Utilizing all annotations more than once is implemented by using more than 1 template image for training. We conduct extensive experiments to train Siamese-DETR with different numbers of template images. As we can see from \Tref{table_results_of_different_training_strategy_swin}, Siamese-DETR achieves the best detection and tracking results when trained with 7 template images. Specifically, compared with the counterpart trained with 1 template image, utilizing 7 template images for training achieves 6.3\% higher mAP@0.5 and 4.1\% higher MOTA. By default, 7 template images are used for the training of Siamese-DETR.

\subsubsection{Tracking-by-Query}

The effectiveness of our simple online tracking pipeline TbQ, has been proved in \Tref{table_comparison_with_different_methods} and \Sref{sec_comparision_of_tracking_performance} by comparing TbQ with other trackers. Here, we further show the effectiveness of the optimized query denoising. Results are shown in \Tref{table_results_of_query_denoising}. As we can see, both the original query denoising~\cite{zhang2022dino} and optimized query denoising are effective to improve the detection and tracking performance. However, as stated in \Sref{sec_dynamic_matching_training_strategy}, the original query denoising introduces some conflicts with the template-image-based object detection/tracking. \revise{With the help of our optimized query denoising, the tracking scenario is more accurately mimicked and the TbQ tracking pipeline is more effectively learned during the training stage, which brings more performance gain than the original query denoising.}

To further study the impact of query denoising, we use ground-truth boxes as query boxes to detect objects. The average confidence score (Avg. Conf.) of predicted boxes and the average IoU (Avg. IoU) between predicted boxes and their corresponding ground-truth boxes are calculated to show the classification and box regression capabilities of Siamese-DETR. As we can see, the original query denoising is negative to the classification capability of Siamese-DETR, and the detection and tracking performance gain mainly comes from the better box regression capability. However, with the optimized query denoising, both the classification and box regression capabilities of Siamese-DETR are promoted.

\subsubsection{Impact of Training Data}
We show the impact of the training data on our Siamese-DETR and open-vocabulary methods (\textit{e.g.}, GLIP-T (B)~\cite{radford2021learning}) in \Tref{table_results_of_different_training_datasets}.

Compared with COCO-trained Siamese-DETR, the LVIS-trained Siamese-DETR achieves poorer detection and tracking performance. We attribute this to the fewer training images in LVIS than COCO (100K vs. 118K). Differently, LVIS-trained GLIP-T (B) performs much better than the COCO-trained one, indicating that the fine-grained category annotations play a key role in boosting performance for open-vocabulary methods. Compared with GLIP-T (B), our Siamese-DETR has a lower demand for category annotations, which reduces the labeling cost while collecting training data. When trained on Objects365, Siamese-DETR is greatly boosted thanks to the larger amount of training data. Such a phenomenon is also observed on GLIP-T (B). However, Siamese-DETR outperforms GLIP-T (B) by a large margin when trained with the same data, demonstrating the effectiveness of our method.

\begin{table*}[!htbp]
\setlength{\tabcolsep}{7.5pt}
\caption{\revise{Results of multi-category multi-object tracking on Urban Tracker~\cite{jodoin2014urban} dataset. The default trackers in Siamese-DETR and OVTrack are adopted. The best results are shown in bold with underline.}}
\label{table_results_of_mcmo}
\centering
\begin{tabular}{c|c|cc|ccccccc} 
\hline
 \multirow{2}*{\revise{Methods}} & \multirow{2}*{\makecell{\revise{Datasets for} \\ \revise{Training}}} &\multicolumn{2}{c|}{\revise{Detection Results}} & \multicolumn{7}{c}{\revise{Tracking Results}} \\
\cline{3-11}
& & \revise{mAP@0.5$\uparrow$}& \revise{mAR$\uparrow$}& \revise{MOTA$\uparrow$}& \revise{IDF1$\uparrow$}& \revise{MT$\uparrow$} & \revise{ML$\downarrow$} & \revise{FP$\downarrow$} & \revise{FN$\downarrow$} &\revise{IDSw$\downarrow$} \\
\hline
\revise{OVTrack~\cite{li2023ovtrack}} &\revise{LVIS~\cite{gupta2019lvis}} &\revise{20.0\%} & \revise{27.4\%}& \revise{5.5\%} & \revise{26.1\%} & \revise{8} & \revise{47} & \revise{\underline{\bf 4110}} & \revise{19309} & \revise{\underline{\bf 68}} \\
\hline
\revise{Siamese-DETR (Ours, Swin-T)} &\revise{COCO~\cite{lin2014microsoft}} &\revise{\underline{\bf 35.9\%}} & \revise{\underline{\bf 30.4\%}}& \revise{\underline{\bf 14.2\%}} & \revise{\underline{\bf 27.2\%}} & \revise{\underline{\bf 11}} & \revise{\underline{\bf 30}}& \revise{4967}& \revise{\underline{\bf 14298}}& \revise{542}  \\
\hline
\end{tabular}
\end{table*}

\begin{figure}
    \scriptsize
    \setlength\tabcolsep{1.2 pt}
    \begin{tabular}{ccccc}
    \centering
    \includegraphics[width=0.19\linewidth]{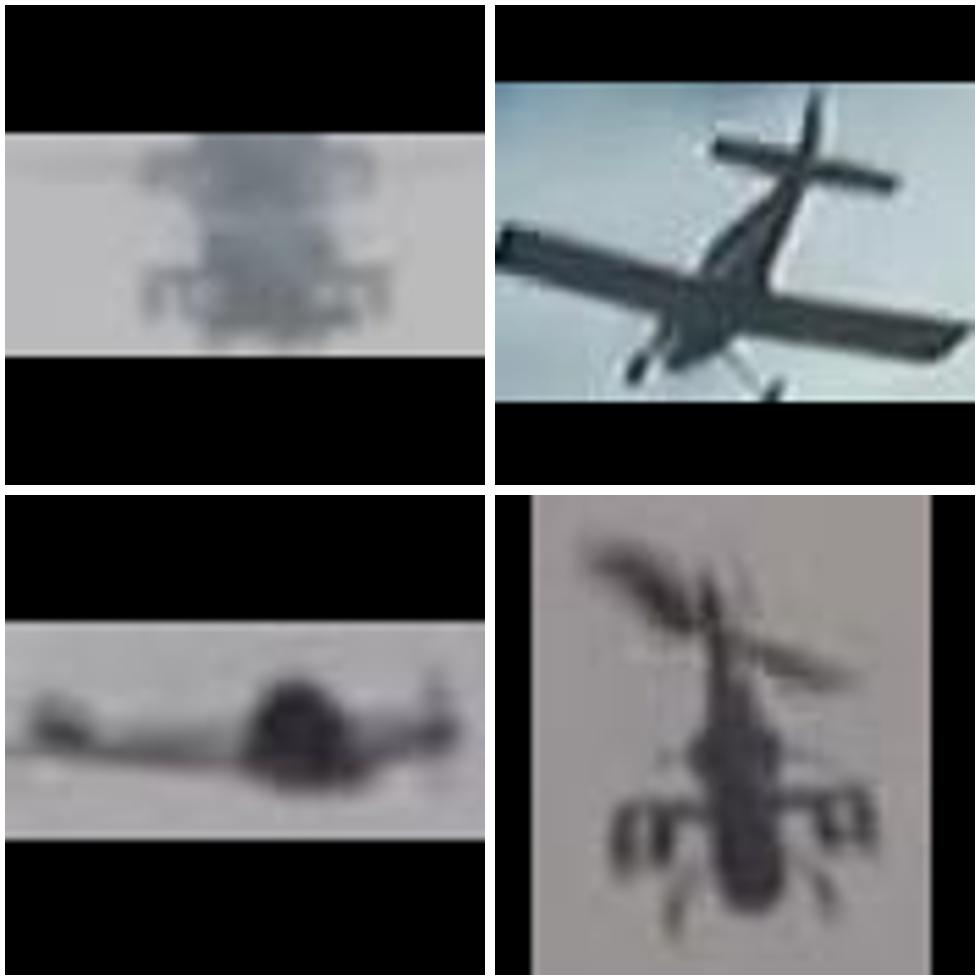} &
    \includegraphics[width=0.19\linewidth]{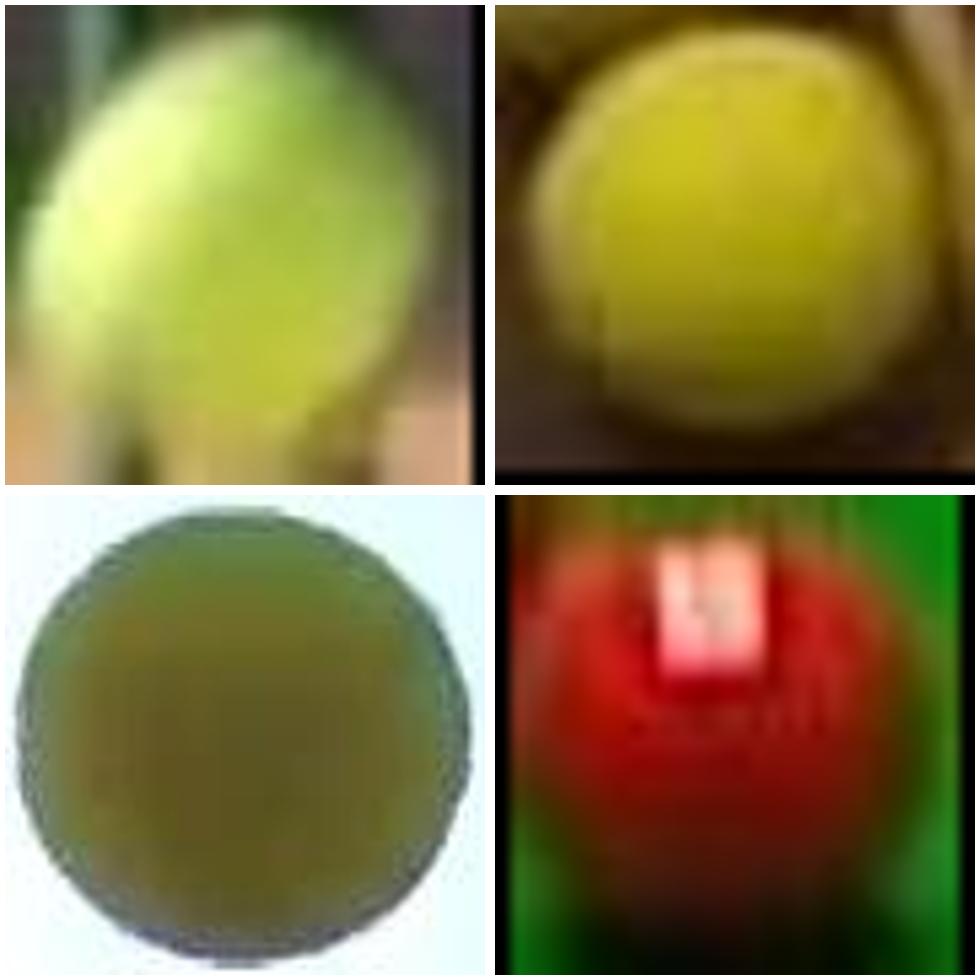} &
    \includegraphics[width=0.19\linewidth]{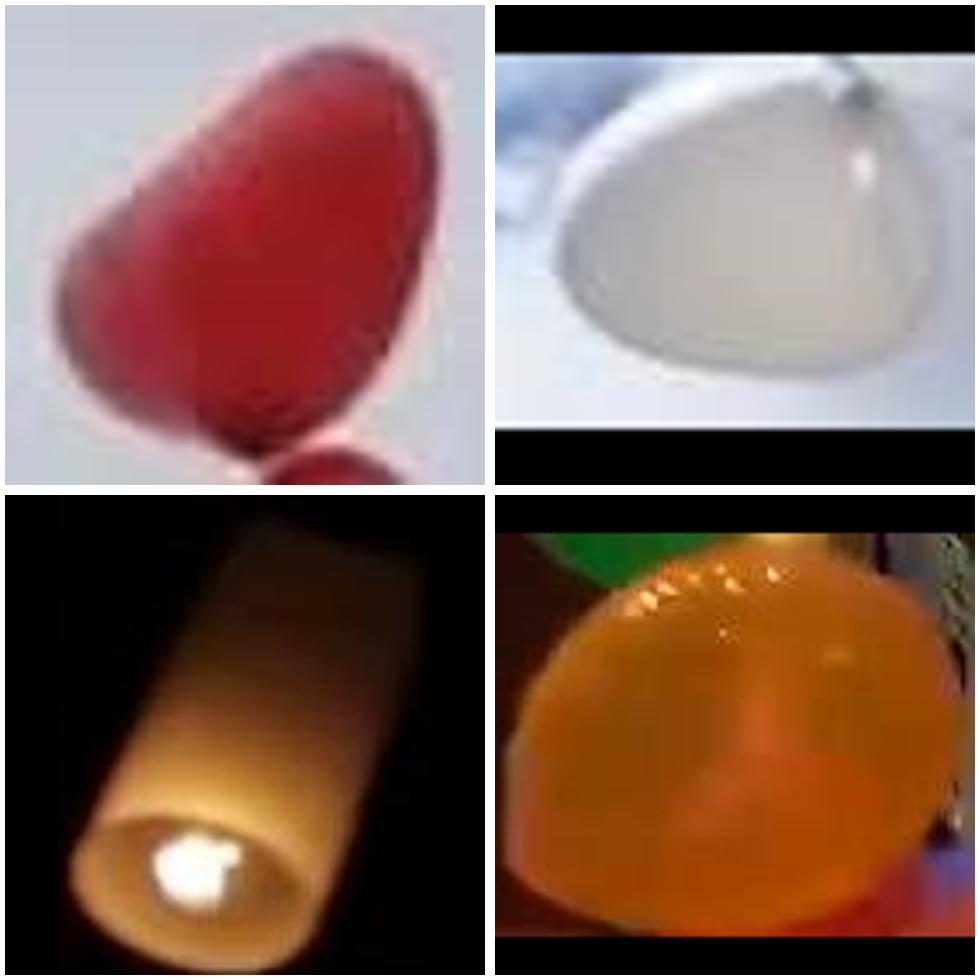} &
    \includegraphics[width=0.19\linewidth]{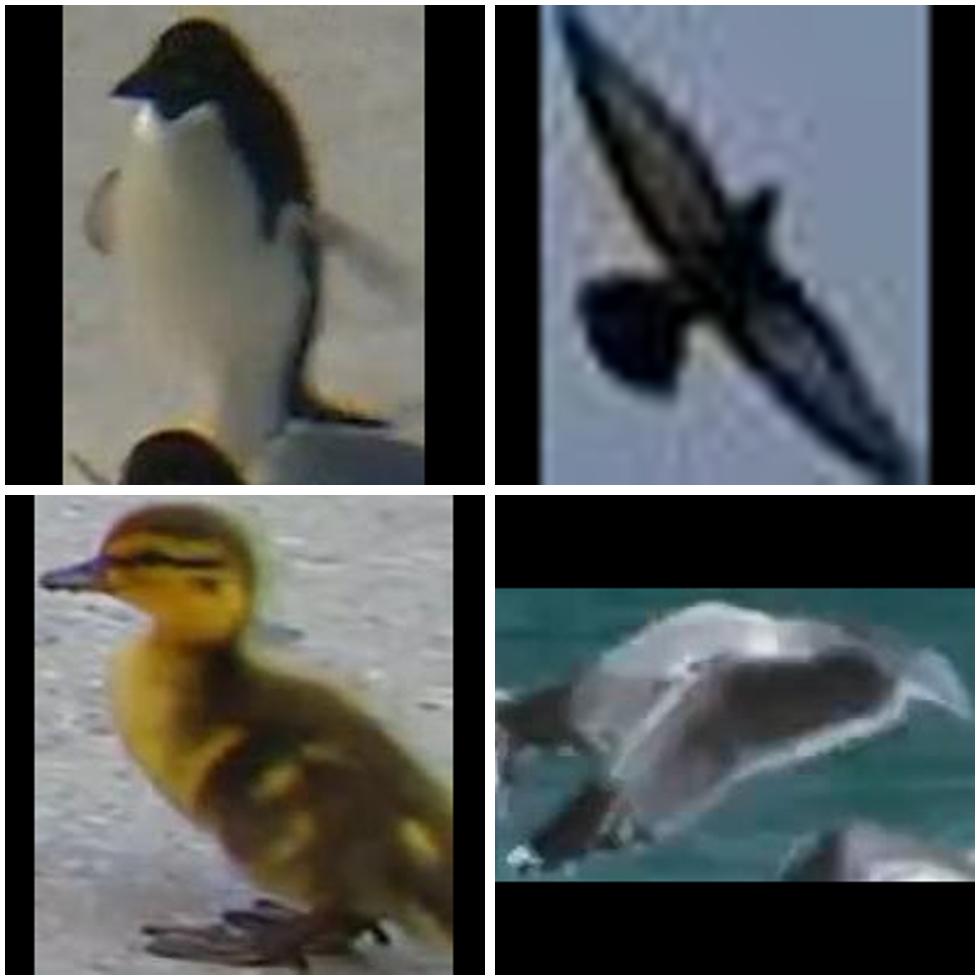} &
    \includegraphics[width=0.19\linewidth]{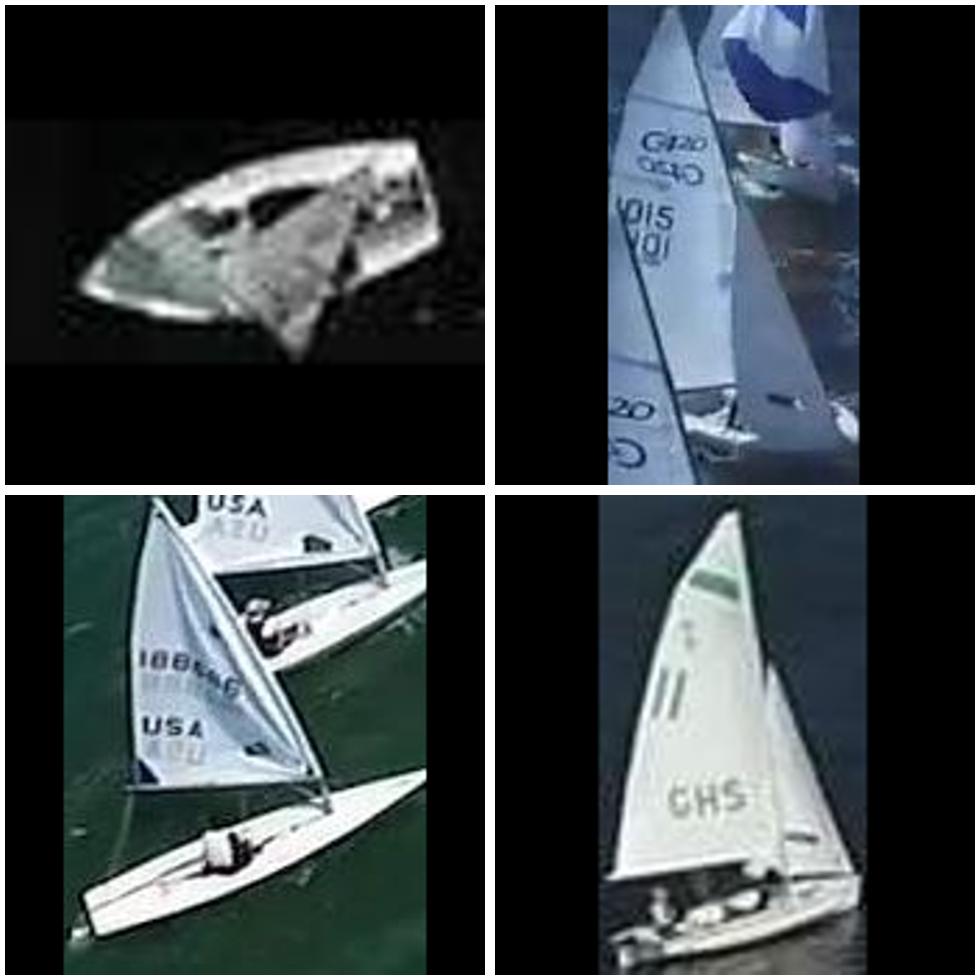}  \\
    \revise{Airplane} & \revise{Ball} & \revise{Balloon} & \revise{Bird} & \revise{Boat} \\
     \includegraphics[width=0.19\linewidth]{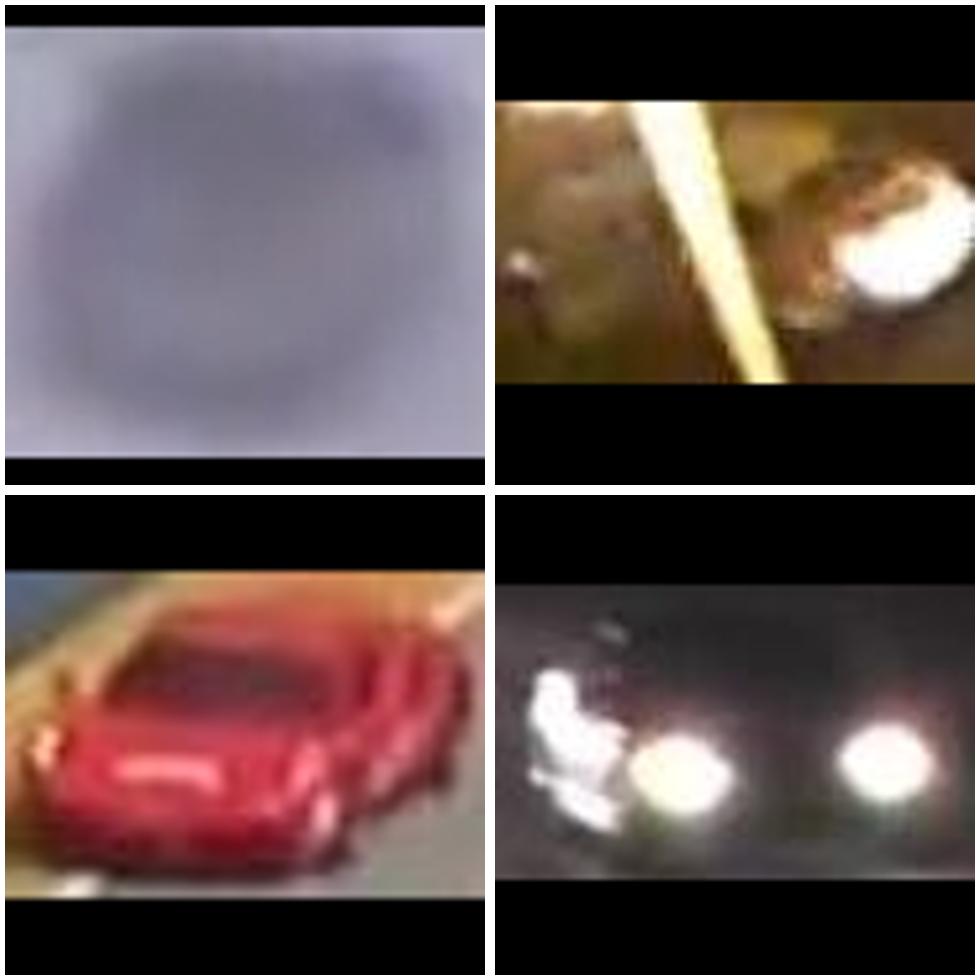} &
    \includegraphics[width=0.19\linewidth]{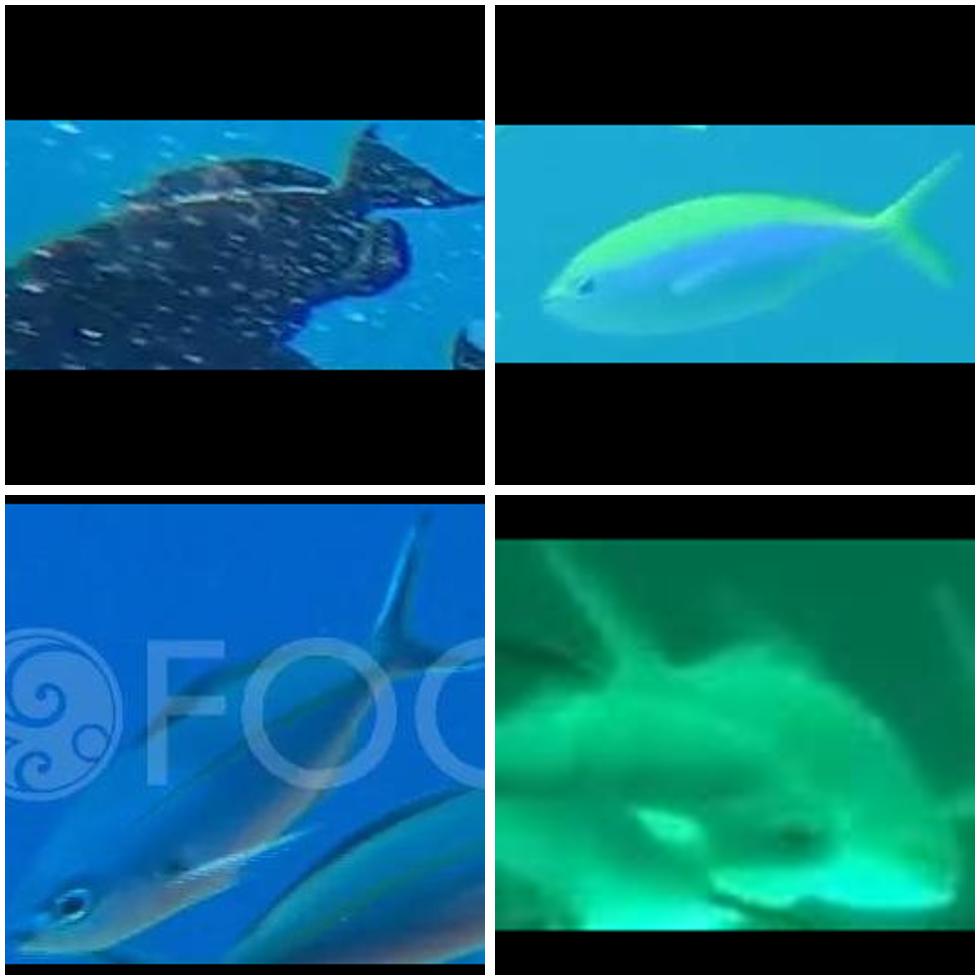} &
    \includegraphics[width=0.19\linewidth]{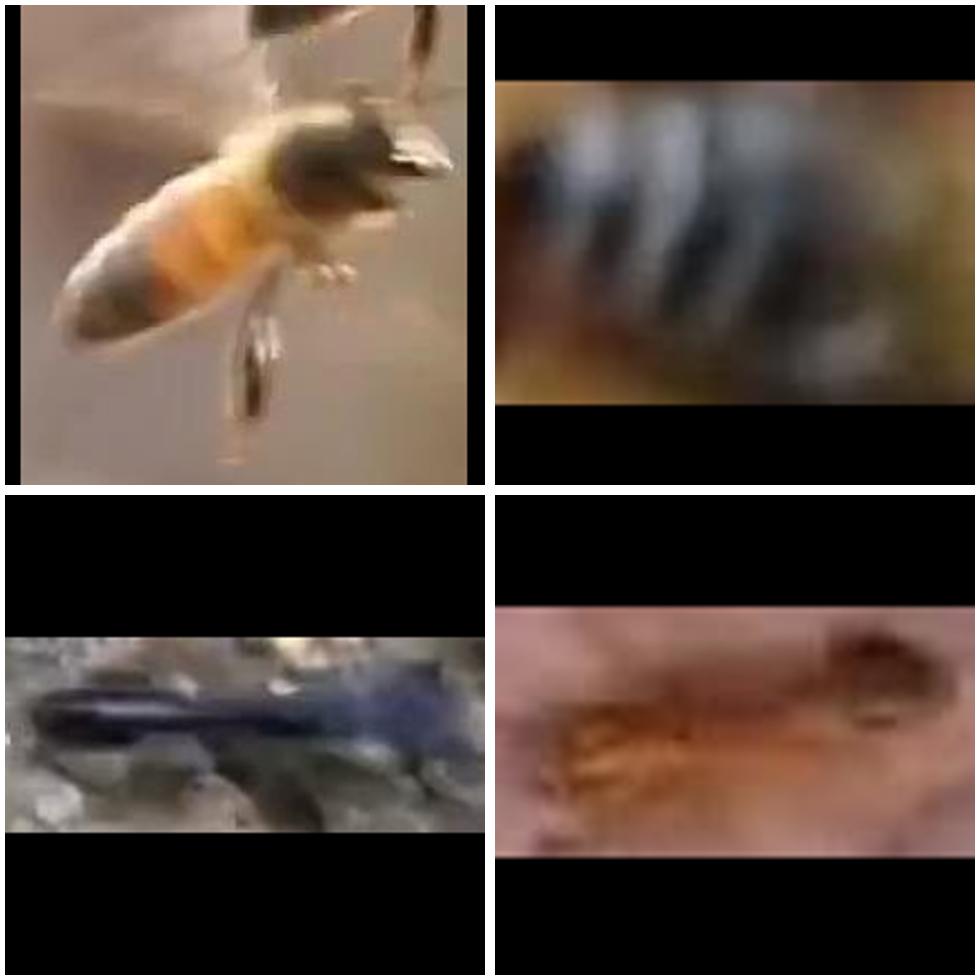} &
    \includegraphics[width=0.19\linewidth]{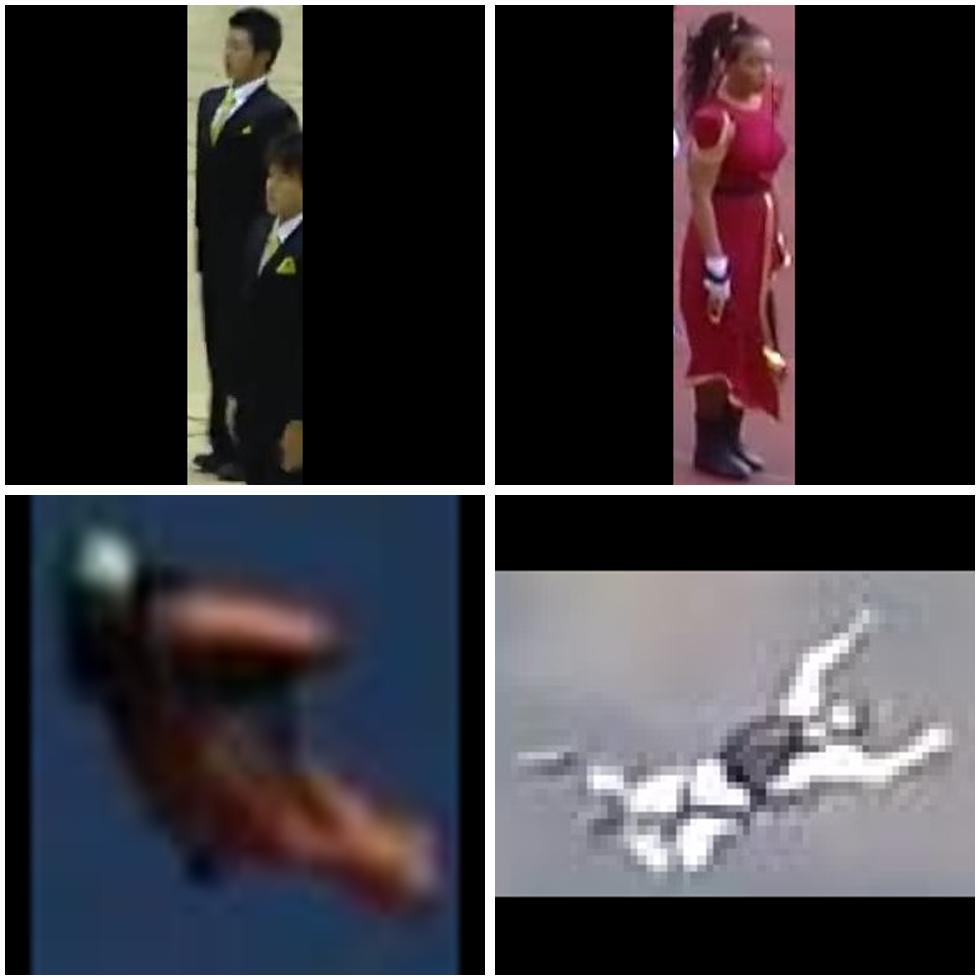} &
    \includegraphics[width=0.19\linewidth]{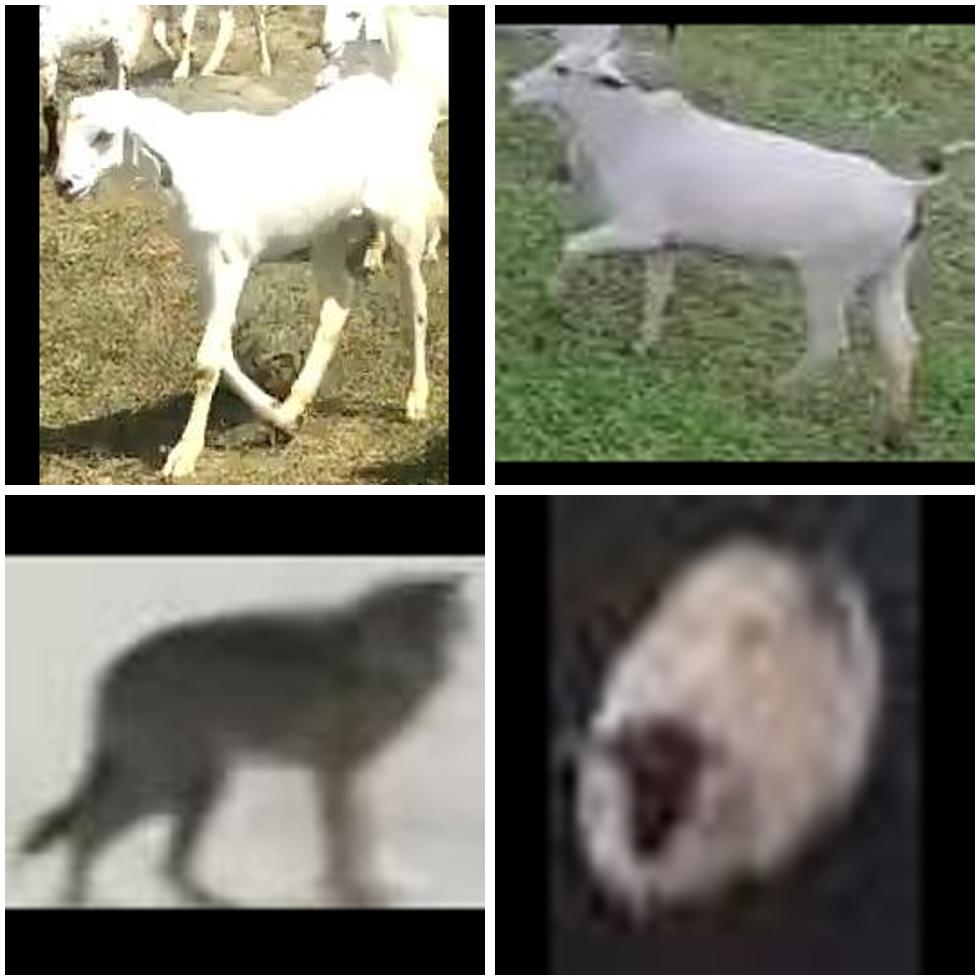}  \\
    \revise{Car} & \revise{Fish} & \revise{Insect} & \revise{Person} & \revise{Stock}
    \end{tabular}
    \caption{\revise{The used template image for each category. We present the used 4 template images for each category since there are 4 videos for each category. All template images are padded to a square resolution.}}
    \label{fig_template_image}
\end{figure}

\begin{table}[!t]
\setlength{\tabcolsep}{17pt}
\caption{\revise{The impact of different sets of template images. Different template images produce different detection performances, but the variabilities are negligible. The Siamese-DETR (Swin-T) trained on COCO is evaluated.}}
\label{table_impact_of_template_images}
\centering
\begin{tabular}{c|cc} 
\hline
\multirow{2}*{\revise{Sets of Template Images}} & \multicolumn{2}{c}{\revise{Detection Results}} \\
\cline{2-3}
& \revise{mAP@50$\uparrow$} &\revise{mAR$\uparrow$} \\
\hline
\revise{First Set (Default)} &\revise{57.5\%} & \revise{46.6\%} \\
\revise{Second Set}& \revise{\underline{\bf 57.7\%}}& \revise{47.2\%} \\
\revise{Third Set}& \revise{\underline{\bf 57.7\%}}& \revise{\underline{\bf 47.4\%}} \\
\hline
\end{tabular}
\end{table}

\subsubsection{\revise{Comparing with DETR-Based MOT Methods}}
\revise{In Section~\ref{sec_compare_with_existing_methods}, several DETR-based methods, including DINO~\cite{zhang2022dino} and Conditional DETR~\cite{meng2021conditional}, are evaluated. However, these methods are mainly designed for object detection, and the tracking results are obtained with different tracking methods. Here, we further compare Siamese-DETR with DETR-based MOT methods (\textit{e.g.}, TrackFormer~\cite{meinhardt2022trackformer}), which are closed-set tracking methods and are mainly designed for pedestrian/person tracking. Taking the consideration that there is no annotated video data for the training of TIMOT methods and TrackFormer needs to be trained on annotated video data, we directly use the publicly available model weight of TrackFormer and compare it with Siamese-DETR on the four person videos in GMOT-40 benchmark. Results are shown in \Tref{table_compare_with_detr_based_mot_methods}. TrackFormer achieves much poorer detection results than Siamese-DETR, resulting a poor tracking results (MOTA $<$ 0 due to the high FP and FN). The reason is that TrackFormer fails to detect the persons in some scenes (\textit{e.g.,} wingsuit flying in the bottom row of the subfigure Person in \Fref{fig_template_image}) and TrackFormer also tends to detect all persons even we focus on standing persons in some scenes (\textit{e.g.,} the up row of the subfigure Person in \Fref{fig_template_image}). Differently, Siamese-DETR can follow the template image to detect/track interested objects in different scenes.}

\begin{figure}
    \centering
    \scriptsize
    \setlength\tabcolsep{2pt}
     \begin{tabular}{ccc}
    \centering
    \includegraphics[width=0.32\linewidth]{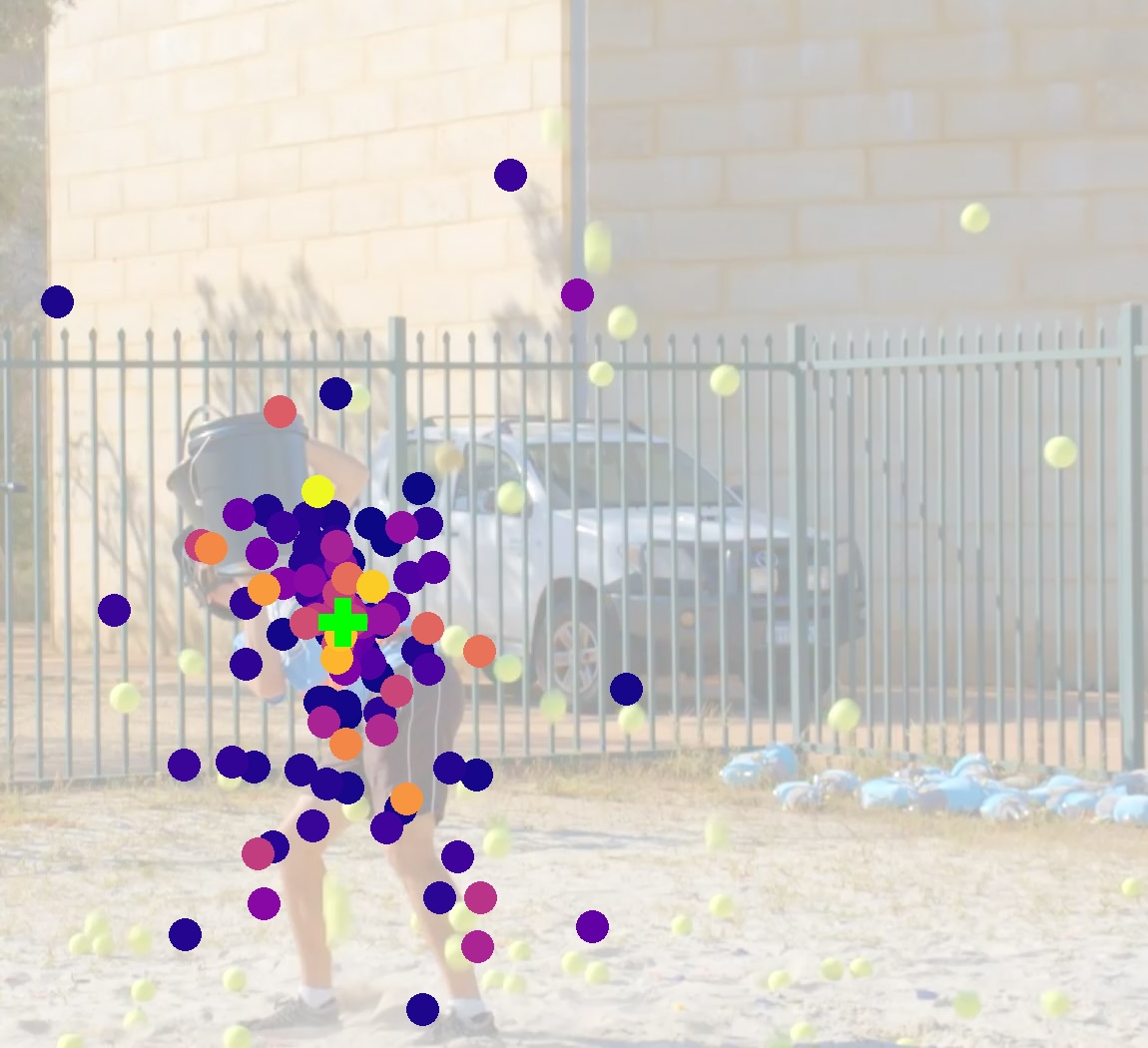} &
    \includegraphics[width=0.32\linewidth]{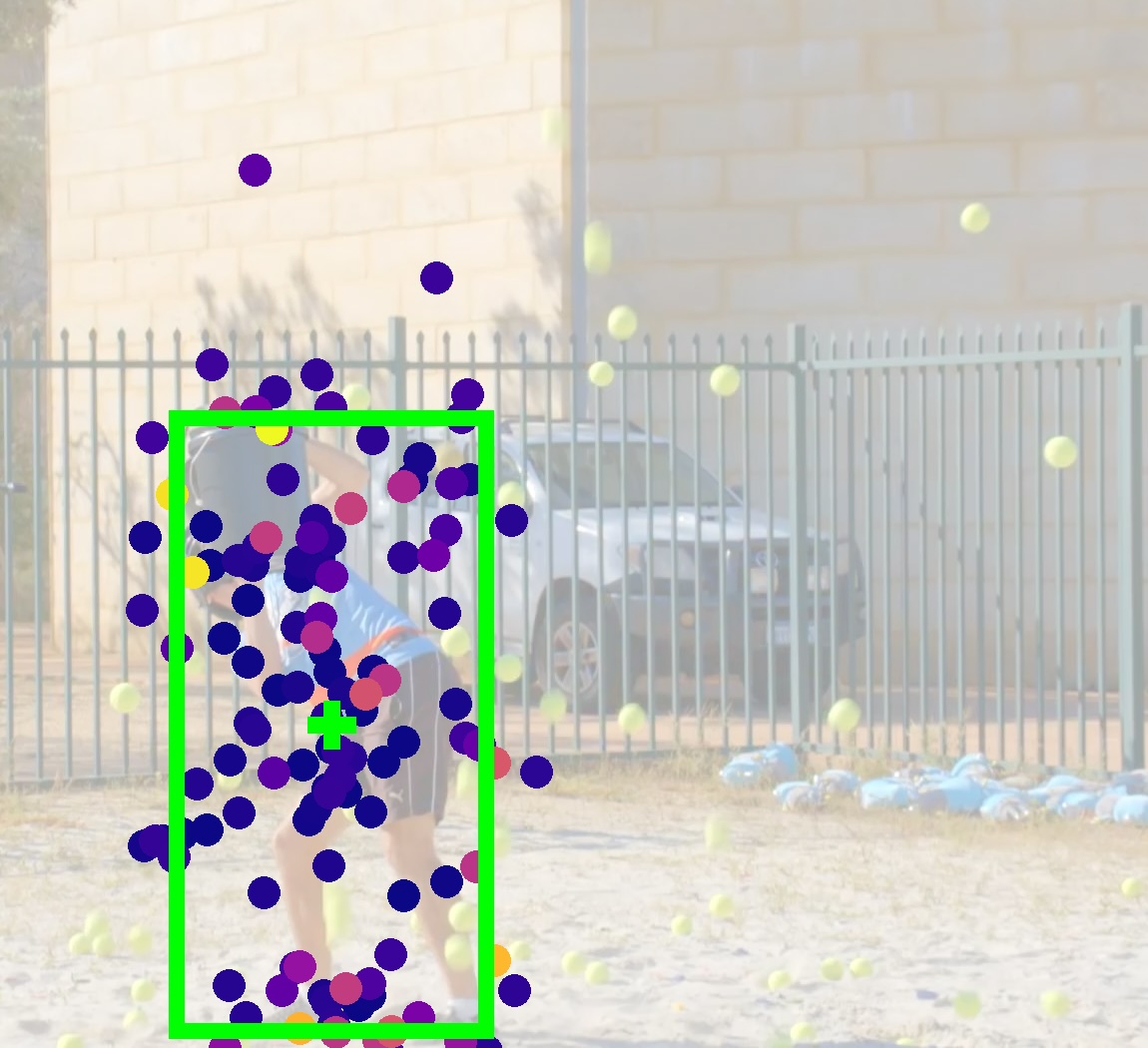} &
    \includegraphics[width=0.32\linewidth]{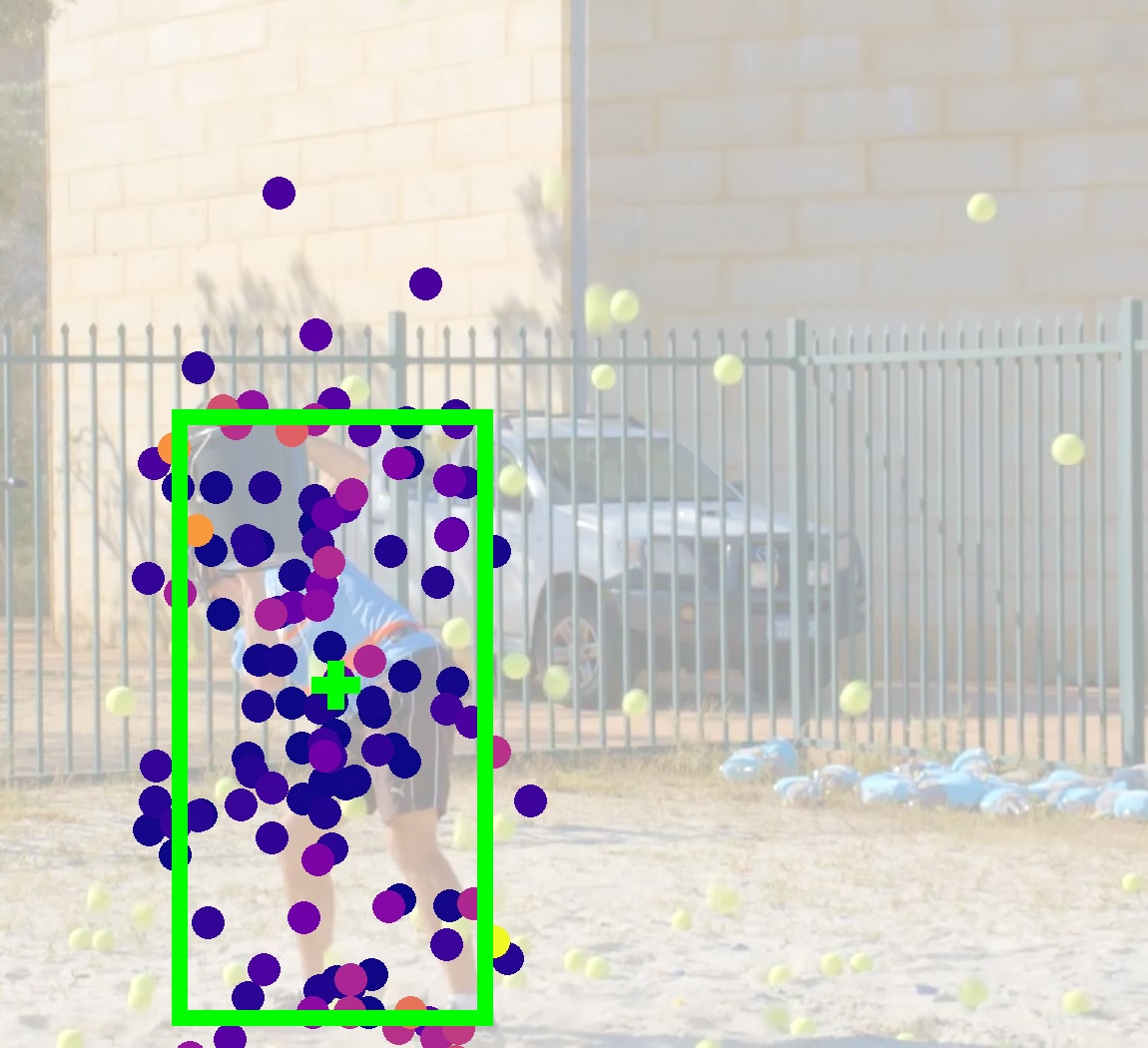}  \\
    \includegraphics[width=0.32\linewidth]{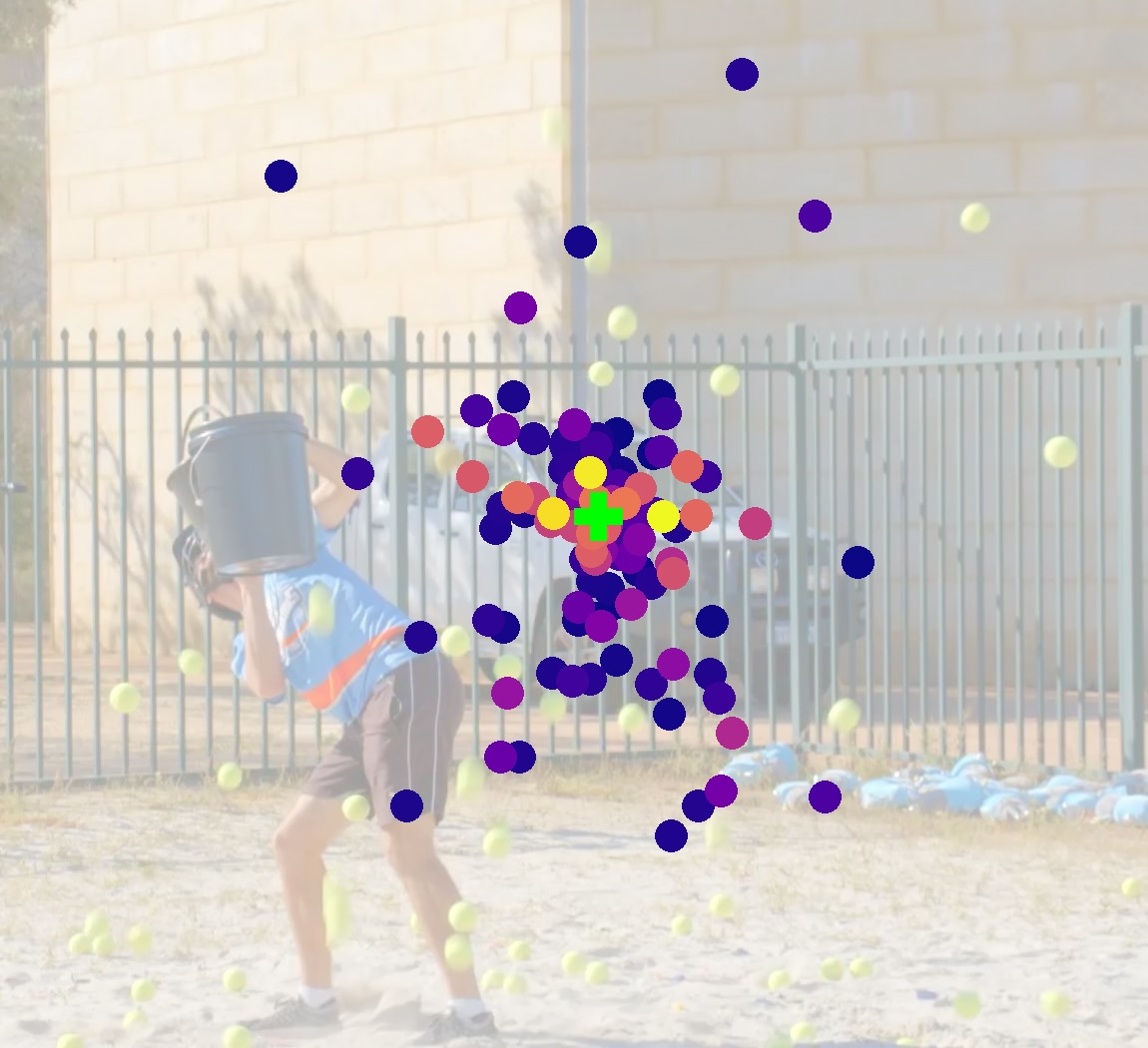} &
    \includegraphics[width=0.32\linewidth]{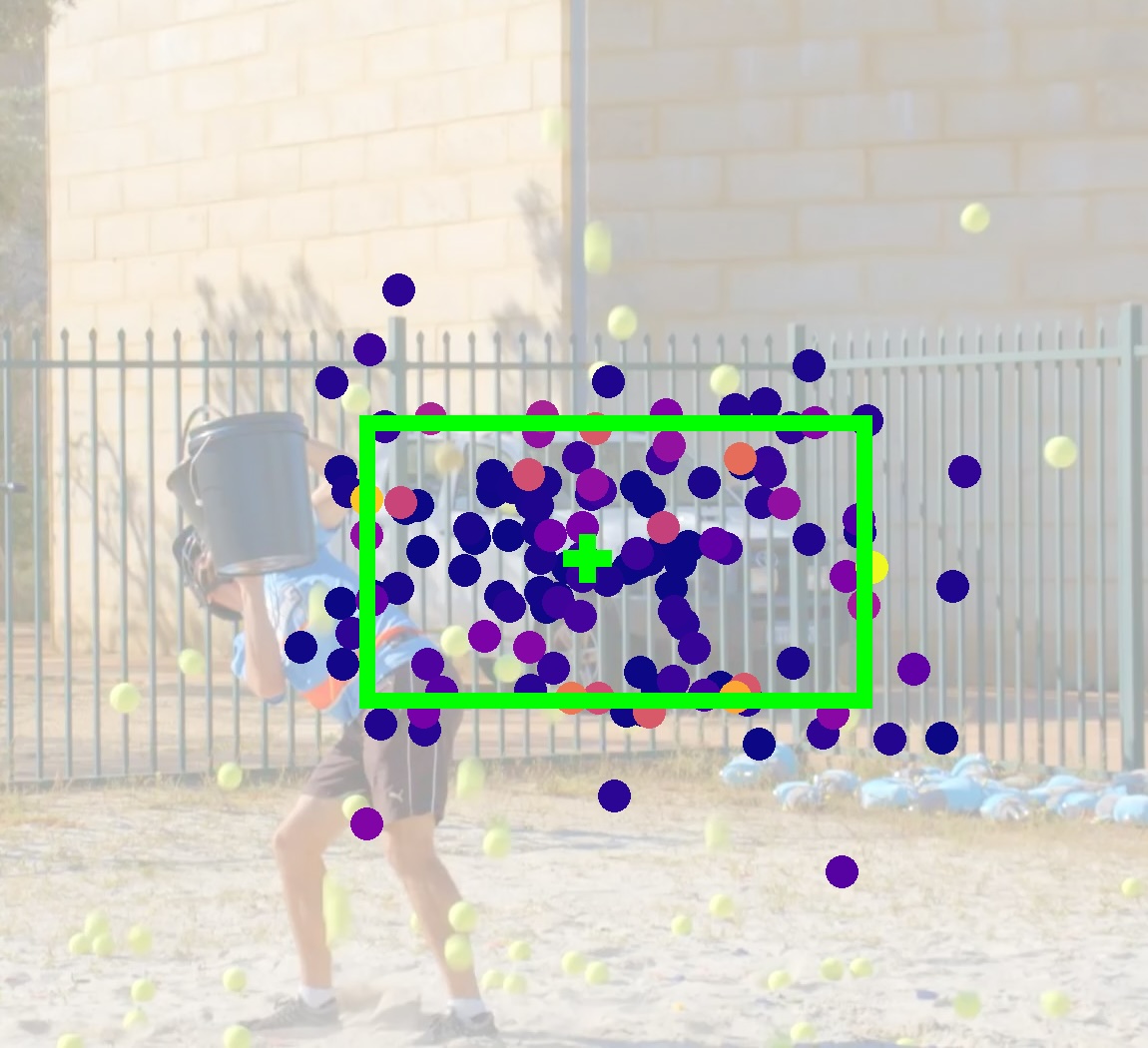} &
    \includegraphics[width=0.32\linewidth]{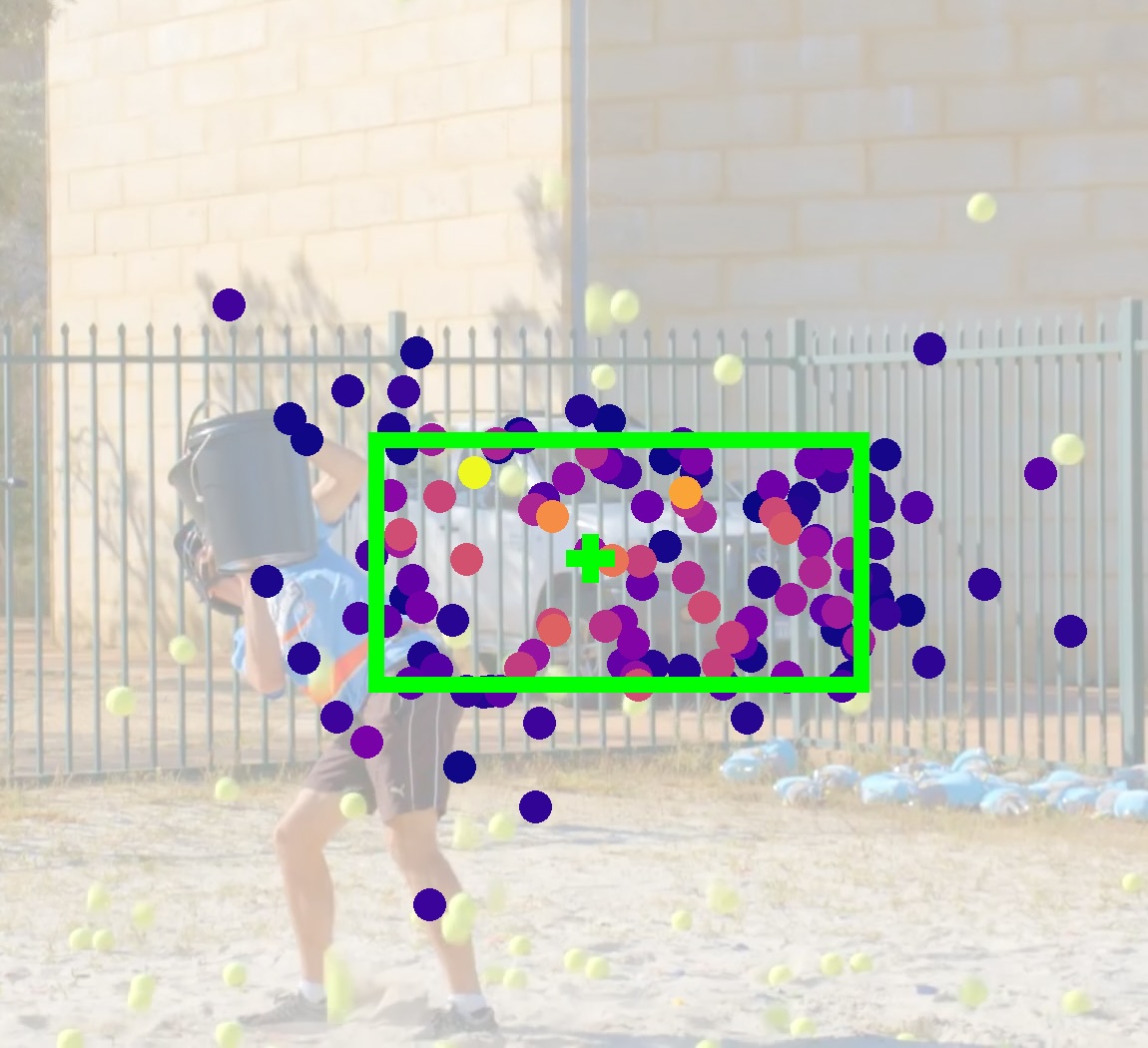} \\
    \multicolumn{3}{c}{\includegraphics[width=1.0\linewidth,height=0.02\linewidth]{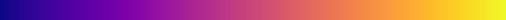}} \\
    \multicolumn{1}{l}{\revise{0.0}} & &\multicolumn{1}{r}{\revise{1.0}} \\
    \end{tabular}
    \caption{\revise{Visualization of attention weights. Left column: the attention weights for different reference points in the last encoder layer. Middle column: the attention weights for different detection object queries (in $Q$) in the last decoder layer. Right column: the attention weights for different tracking object queries (in $\hat{Q}$) in the last decoder layer. The reference points are shown with cross markers. The sampling points are marked as filled circles with the attention weights encoded with different colors. The rectangle boxes are the predicted boxes by the decoder.}}
    \label{fig_attention_weights}
\end{figure}

\subsubsection{\revise{Multi-Category Multi-Object Tracking}}
\revise{Our Siamese-DETR has the capability to track different categories of objects by providing a template image for each category for the reason that it supports multiple template images simultaneously. For better comparison, OVTrack~\cite{li2023ovtrack} is also evaluated by providing all category names to it. As we can see, Siamese-DETR achieves much better detection and tracking performance than OVTrack. The poor performance of OVTrack mainly comes from the higher FN. Through visualization, we find that OVTrack fails to detect some objects if the provided text prompt cannot describe them in detail. However, providing fine-grained text prompts for all objects is impractical in real applications.}

\subsubsection{\revise{Impact of Template Images}}
\revise{The default used template images for each video are presented in \Fref{fig_template_image}. It can be seen that there exists variability within the same category and the impact of different template images needs to be discussed. To do this, we further randomly sample another two sets of template images. The detection results on GMOT-40 are shown in \Tref{table_impact_of_template_images}. As we can see, though different sets of template images produce different detection performances, the variabilities are negligible, demonstrating the generalization ability of Siamese-DETR on template images. A carefully selected set of template images may produce better performance, but it is not our intention.}

\begin{figure}
    \centering
    \scriptsize
    \setlength\tabcolsep{0.8pt}
    \begin{tabular}{cccc}
    \centering
    & Frame 001 & Frame 011 & Frame 021\\
    \rotatebox{90}{\revise{Ground-Truth}} &\includegraphics[width=0.32\linewidth]{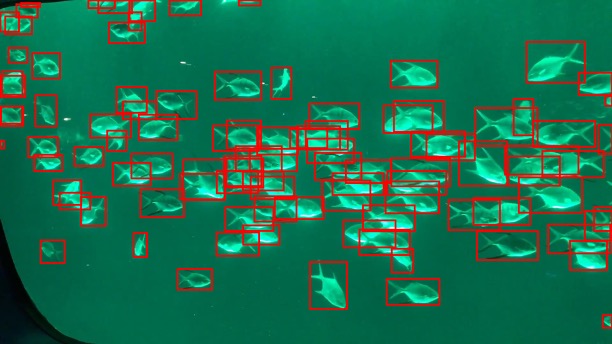} &
    \includegraphics[width=0.32\linewidth]{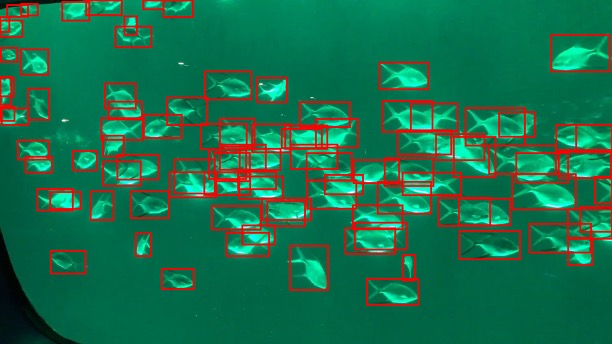} &
    \includegraphics[width=0.32\linewidth]{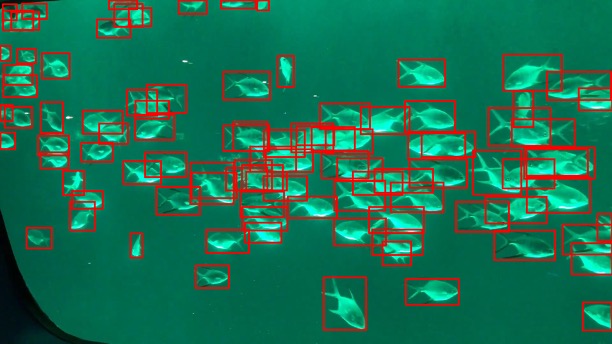} \\
    \rotatebox{90}{\revise{\quad Prediction}} &\includegraphics[width=0.32\linewidth]{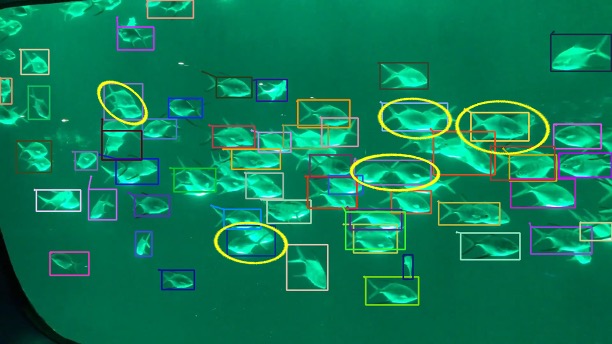} &
    \includegraphics[width=0.32\linewidth]{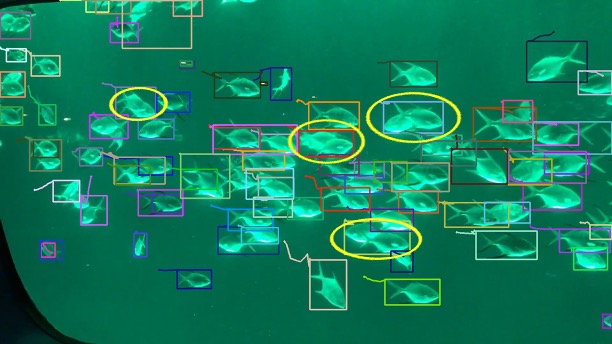} &
    \includegraphics[width=0.32\linewidth]{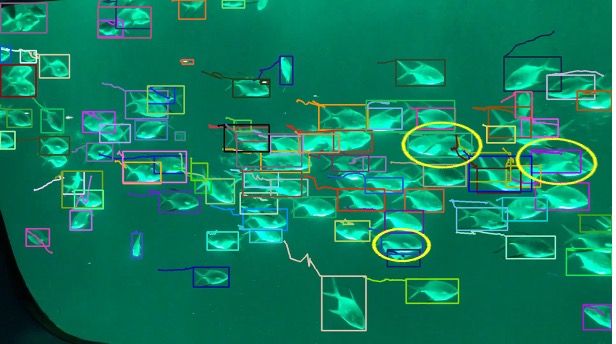} \\
    \hline
    
    & Frame 081 & Frame 091 & Frame 101\\
    \rotatebox{90}{\revise{Ground-Truth}} &\includegraphics[width=0.32\linewidth]{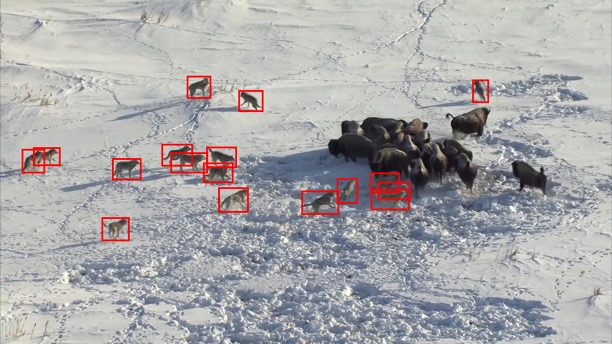} &
    \includegraphics[width=0.32\linewidth]{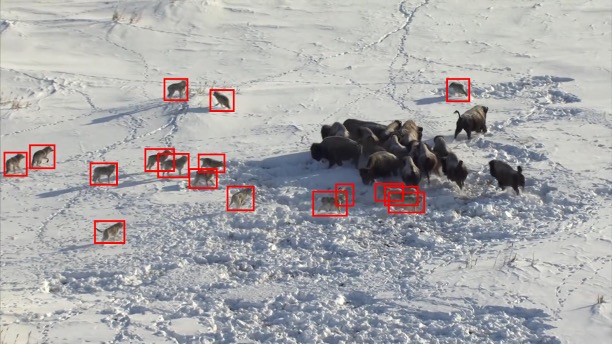} &
    \includegraphics[width=0.32\linewidth]{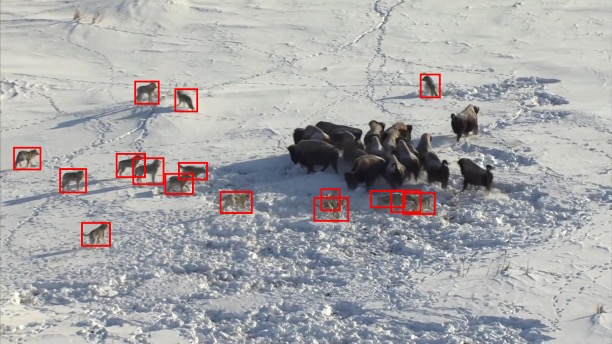} \\
    \rotatebox{90}{\revise{\quad Prediction}} &\includegraphics[width=0.32\linewidth]{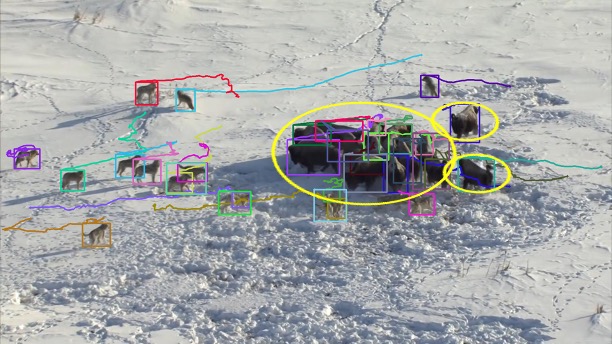} &
    \includegraphics[width=0.32\linewidth]{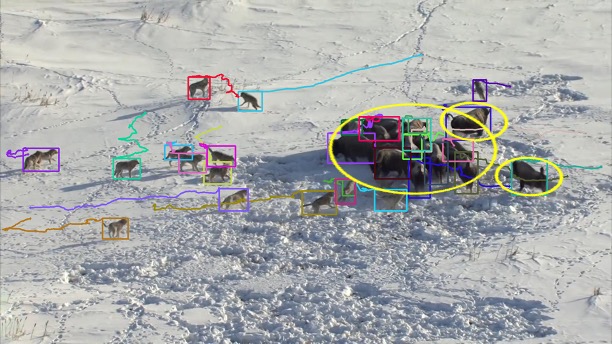} &
    \includegraphics[width=0.32\linewidth]{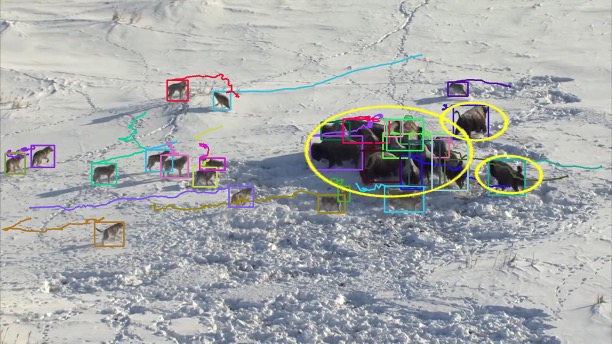} \\
    \end{tabular}
    \caption{\revise{Two failure cases of Siamese-DETR. Siamese-DETR fails to detect/track objects when they are highly overlapped with each other (the up case) and produces some false positives if some objects share a similar appearance with the interested objects (the bottom case). For each case, the ground-truth boxes are plotted (red boxes) for reference. Please pay attention to the objects within the ellipses.}}
    \label{fig_failure_cases}
\end{figure}

\subsubsection{\revise{Visualization of Attention Weights}}
\revise{To show the effectiveness of Siamese-DETR, we show some attention weights produced by transforemr encoder and decoder layers. For the reason that Siamese-DETR is implemented based on DINO~\cite{zhang2022dino} and Deformable-DETR~\cite{zhu2020deformable}, where the attention is calculated in a sparse manner, we first select a reference point and then show the attention weights of all sampled points that contribute to this reference point. Results are shown in \Fref{fig_attention_weights}. For the attention weights produced by the encoder layer (left column), the reference points mainly attend to foreground sampling points. For the attention weights produced by the decoder layer (middle and right columns), the reference points mainly attend to the sampling points in object extremities. From the visualization results in the right column, we can see that the objects can successfully draw attention from the model based on the tracked boxes in the previous frame.}

\subsubsection{\revise{Failure Cases}}
\revise{The two typical failure cases are presented in \Fref{fig_failure_cases}. For the up case, Siamese-DETR fails to detect/track different object instances if they are heavily overlapped with each other. For the bottom case, Siamese-DETR produces some false positive boxes when some objects share a similar appearance with the interested objects.}

\section{Conclusions and Limitations}
\label{sec_conclusion}
In this paper, we focus on \revise{template-image-based multi-object tracking}, where the interested objects are described by the given template image. We take advantage of object queries in DETR variants and propose Siamese-DETR \revise{to track generic multi-objects}.  In order to detect different scales of objects that share the same category with the given template image, multi-scale object queries are designed, where the query contents are obtained from the template image. In addition, a dynamic matching training strategy is proposed to train Siamese-DETR efficiently on commonly used detection datasets. To handle the \revise{scarcity} of video training data, the query denoising is adopted and optimized, which mimics the tracking scenarios on static images.  While tracking online, the tracking pipeline is simplified by incorporating the tracked boxes as additional query boxes. Object detection and tracking are performed simultaneously and the complex data association is replaced with the simpler NMS operation. Experimental results demonstrate the effectiveness of the proposed Siamese-DETR.

The main limitations of Siamese-DETR lie in twofold: 1) Siamese-DETR is trained with a \textit{two-category} detection task, where the objects are classified into positive and negative samples. An object may be treated as a positive sample for different template images if they share a similar appearance. This may be mitigated by providing several different template images and training the model with \textit{multi-category} detection task; (2) Siamese-DETR tracks objects solely based on the tracked boxes in the previous frame without the exploration of appearance cues. The absence of appearance cues may result in tracking failure when occlusion between different objects happens, producing a higher IDSw and lower IDF1. This can be solved by pairing the tracked boxes with the corresponding appearance features rather than the features extracted from the template image. We leave these limitations to our future works.

\section*{Acknowledgments}
This work was supported by the National Key R\&D Program of China (2022YFC3300704), the National Natural Science Foundation of China (62331006, 62171038, and 62088101), and the Fundamental Research Funds for the Central Universities.

\bibliographystyle{IEEEtran}
\bibliography{cite.bib}

\begin{thebibliography}{10}
\providecommand{\url}[1]{#1}
\csname url@samestyle\endcsname
\providecommand{\newblock}{\relax}
\providecommand{\bibinfo}[2]{#2}
\providecommand{\BIBentrySTDinterwordspacing}{\spaceskip=0pt\relax}
\providecommand{\BIBentryALTinterwordstretchfactor}{4}
\providecommand{\BIBentryALTinterwordspacing}{\spaceskip=\fontdimen2\font plus
\BIBentryALTinterwordstretchfactor\fontdimen3\font minus \fontdimen4\font\relax}
\providecommand{\BIBforeignlanguage}[2]{{%
\expandafter\ifx\csname l@#1\endcsname\relax
\typeout{** WARNING: IEEEtran.bst: No hyphenation pattern has been}%
\typeout{** loaded for the language `#1'. Using the pattern for}%
\typeout{** the default language instead.}%
\else
\language=\csname l@#1\endcsname
\fi
#2}}
\providecommand{\BIBdecl}{\relax}
\BIBdecl

\bibitem{bergmann2019tracking}
P.~Bergmann, T.~Meinhardt, and L.~Leal-Taixe, ``Tracking without bells and whistles,'' in \emph{Proceedings of the IEEE/CVF International Conference on Computer Vision}, 2019, pp. 941--951.

\bibitem{braso2020learning}
G.~Bras{\'o} and L.~Leal-Taix{\'e}, ``Learning a neural solver for multiple object tracking,'' in \emph{Proceedings of the IEEE/CVF Conference on Computer Vision and Pattern Recognition}, 2020, pp. 6247--6257.

\bibitem{wojke2017simple}
N.~Wojke, A.~Bewley, and D.~Paulus, ``Simple online and realtime tracking with a deep association metric,'' in \emph{Proceedings of IEEE International Conference on Image Processing}, 2017, pp. 3645--3649.

\bibitem{chu2019famnet}
P.~Chu and H.~Ling, ``Famnet: Joint learning of feature, affinity and multi-dimensional assignment for online multiple object tracking,'' in \emph{Proceedings of the IEEE/CVF International Conference on Computer Vision}, 2019, pp. 6172--6181.

\bibitem{zhang2022bytetrack}
Y.~Zhang, P.~Sun, Y.~Jiang, D.~Yu, F.~Weng, Z.~Yuan, P.~Luo, W.~Liu, and X.~Wang, ``Bytetrack: Multi-object tracking by associating every detection box,'' in \emph{Proceedings of the European Conference on Computer Vision}, 2022, pp. 1--21.

\bibitem{zhou2020tracking}
X.~Zhou, V.~Koltun, and P.~Kr{\'a}henb{\'u}hl, ``Tracking objects as points,'' in \emph{Proceedings of the European Conference on Computer Vision}, 2020, pp. 474--490.

\bibitem{li2023ovtrack}
S.~Li, T.~Fischer, L.~Ke, H.~Ding, M.~Danelljan, and F.~Yu, ``Ovtrack: Open-vocabulary multiple object tracking,'' in \emph{Proceedings of the IEEE/CVF Conference on Computer Vision and Pattern Recognition}, 2023, pp. 5567--5577.

\bibitem{bai2021gmot}
H.~Bai, W.~Cheng, P.~Chu, J.~Liu, K.~Zhang, and H.~Ling, ``Gmot-40: A benchmark for generic multiple object tracking,'' in \emph{Proceedings of the IEEE/CVF Conference on Computer Vision and Pattern Recognition}, 2021, pp. 6719--6728.

\bibitem{devlin2018bert}
J.~D. M.-W.~C. Kenton and L.~K. Toutanova, ``Bert: Pre-training of deep bidirectional transformers for language understanding,'' in \emph{Proceedings of the North American Chapter of the Association for Computational Linguistics: Human Language Technologies}, 2019, pp. 4171--4186.

\bibitem{radford2021learning}
A.~Radford, J.~W. Kim, C.~Hallacy, A.~Ramesh, G.~Goh, S.~Agarwal, G.~Sastry, A.~Askell, P.~Mishkin, J.~Clark \emph{et~al.}, ``Learning transferable visual models from natural language supervision,'' in \emph{Proceedings of the IEEE International Conference on Machine Learning}, 2021, pp. 8748--8763.

\bibitem{lin2014microsoft}
T.-Y. Lin, M.~Maire, S.~Belongie, J.~Hays, P.~Perona, D.~Ramanan, P.~Doll{\'a}r, and C.~L. Zitnick, ``Microsoft coco: Common objects in context,'' in \emph{Proceedings of the European Conference on Computer Vision}, 2014, pp. 740--755.

\bibitem{luo2013generic}
W.~Luo and T.-K. Kim, ``Generic object crowd tracking by multi-task learning.'' in \emph{Proceedings of the British Machine Vision Conference}, vol.~1, no.~2, 2013, p.~3.

\bibitem{luo2014bi}
W.~Luo, T.-K. Kim, B.~Stenger, X.~Zhao, and R.~Cipolla, ``Bi-label propagation for generic multiple object tracking,'' in \emph{Proceedings of the IEEE/CVF Conference on Computer Vision and Pattern Recognition}, 2014, pp. 1290--1297.

\bibitem{evgeniou2004regularized}
T.~Evgeniou and M.~Pontil, ``Regularized multi--task learning,'' in \emph{Proceedings of the tenth ACM SIGKDD International Conference on Knowledge Discovery and Data Mining}, 2004, pp. 109--117.

\bibitem{dalal2005histograms}
N.~Dalal and B.~Triggs, ``Histograms of oriented gradients for human detection,'' in \emph{Proceedings of the IEEE/CVF Conference on Computer Vision and Pattern Recognition}, 2005, pp. 886--893.

\bibitem{fu2023category}
Y.~Fu, H.~Liu, Y.~Zou, S.~Wang, Z.~Li, and D.~Zheng, ``Category-level band learning based feature extraction for hyperspectral image classification,'' \emph{IEEE Transactions on Geoscience and Remote Sensing}, pp. 1--16, 2023.

\bibitem{huang2020globaltrack}
L.~Huang, X.~Zhao, and K.~Huang, ``Globaltrack: A simple and strong baseline for long-term tracking,'' in \emph{Proceedings of the AAAI Conference on Artificial Intelligence}, vol.~34, no.~07, 2020, pp. 11\,037--11\,044.

\bibitem{fan2021lasot}
H.~Fan, H.~Bai, L.~Lin, F.~Yang, P.~Chu, G.~Deng, S.~Yu, M.~Huang, J.~Liu, Y.~Xu \emph{et~al.}, ``Lasot: A high-quality large-scale single object tracking benchmark,'' \emph{International Journal of Computer Vision}, vol. 129, pp. 439--461, 2021.

\bibitem{huang2019got}
L.~Huang, X.~Zhao, and K.~Huang, ``Got-10k: A large high-diversity benchmark for generic object tracking in the wild,'' \emph{IEEE Transactions on Pattern Analysis and Machine Intelligence}, vol.~43, no.~5, pp. 1562--1577, 2019.

\bibitem{bewley2016simple}
A.~Bewley, Z.~Ge, L.~Ott, F.~Ramos, and B.~Upcroft, ``Simple online and realtime tracking,'' in \emph{Proceedings of the IEEE International Conference on Image Processing}, 2016, pp. 3464--3468.

\bibitem{bochinski2017high}
E.~Bochinski, V.~Eiselein, and T.~Sikora, ``High-speed tracking-by-detection without using image information,'' in \emph{Proceedings of the IEEE international conference on advanced video and signal based surveillance}, 2017, pp. 1--6.

\bibitem{zhang2022dino}
H.~Zhang, F.~Li, S.~Liu, L.~Zhang, H.~Su, J.~Zhu, L.~Ni, and H.-Y. Shum, ``Dino: Detr with improved denoising anchor boxes for end-to-end object detection,'' in \emph{The Eleventh International Conference on Learning Representations}, 2022.

\bibitem{zhu2020deformable}
X.~Zhu, W.~Su, L.~Lu, B.~Li, X.~Wang, and J.~Dai, ``Deformable detr: Deformable transformers for end-to-end object detection,'' in \emph{International Conference on Learning Representations}, 2020.

\bibitem{li2022dn}
F.~Li, H.~Zhang, S.~Liu, J.~Guo, L.~M. Ni, and L.~Zhang, ``Dn-detr: Accelerate detr training by introducing query denoising,'' in \emph{Proceedings of the IEEE/CVF Conference on Computer Vision and Pattern Recognition}, 2022, pp. 13\,619--13\,627.

\bibitem{liu2022dab}
S.~Liu, F.~Li, H.~Zhang, X.~Yang, X.~Qi, H.~Su, J.~Zhu, and L.~Zhang, ``Dab-detr: Dynamic anchor boxes are better queries for detr,'' in \emph{International Conference on Learning Representations}, 2021.

\bibitem{li2024supervise}
M.~Li, Y.~Fu, T.~Zhang, and G.~Wen, ``Supervise-assisted self-supervised deep-learning method for hyperspectral image restoration,'' \emph{IEEE Transactions on Neural Networks and Learning Systems}, pp. 1--14, 2024.

\bibitem{roshan2012gmcp}
A.~Roshan~Zamir, A.~Dehghan, and M.~Shah, ``Gmcp-tracker: Global multi-object tracking using generalized minimum clique graphs,'' in \emph{Proceedings of the European Conference on Computer Vision}, 2012, pp. 343--356.

\bibitem{zhang2008global}
L.~Zhang, Y.~Li, and R.~Nevatia, ``Global data association for multi-object tracking using network flows,'' in \emph{Proceedings of the IEEE/CVF Conference on Computer Vision and Pattern Recognition}, 2008, pp. 1--8.

\bibitem{keuper2015efficient}
M.~Keuper, E.~Levinkov, N.~Bonneel, G.~Lavou{\'e}, T.~Brox, and B.~Andres, ``Efficient decomposition of image and mesh graphs by lifted multicuts,'' in \emph{Proceedings of the IEEE International Conference on Computer Vision}, 2015, pp. 1751--1759.

\bibitem{tang2017multiple}
S.~Tang, M.~Andriluka, B.~Andres, and B.~Schiele, ``Multiple people tracking by lifted multicut and person re-identification,'' in \emph{Proceedings of the IEEE Conference on Computer Vision and Pattern Recognition}, 2017, pp. 3539--3548.

\bibitem{chen2023instance}
L.~Chen, Y.~Fu, K.~Wei, D.~Zheng, and F.~Heide, ``Instance segmentation in the dark,'' \emph{International Journal of Computer Vision}, vol. 131, no.~8, pp. 2198--2218, 2023.

\bibitem{fu2023raw}
Y.~Fu, Y.~Hong, Y.~Zou, Q.~Liu, Y.~Zhang, N.~Liu, and C.~Yan, ``Raw image based over-exposure correction using channel-guidance strategy,'' \emph{IEEE Transactions on Circuits and Systems for Video Technology}, vol.~34, pp. 2749--2762, 2023.

\bibitem{fang2018recurrent}
K.~Fang, Y.~Xiang, X.~Li, and S.~Savarese, ``Recurrent autoregressive networks for online multi-object tracking,'' in \emph{Proceedings of IEEE Winter Conference on Applications of Computer Vision}, 2018, pp. 466--475.

\bibitem{zhang2024deep}
T.~Zhang, Y.~Fu, J.~Zhang, and C.~Yan, ``Deep guided attention network for joint denoising and demosaicing in real image,'' \emph{Chinese Journal of Electronics}, vol.~33, no.~1, pp. 303--312, 2024.

\bibitem{ren2020tracking}
W.~Ren, X.~Wang, J.~Tian, Y.~Tang, and A.~B. Chan, ``Tracking-by-counting: Using network flows on crowd density maps for tracking multiple targets,'' \emph{IEEE Transactions on Image Processing}, vol.~30, pp. 1439--1452, 2020.

\bibitem{lee2023decode}
S.-H. Lee, D.-H. Park, and S.-H. Bae, ``Decode-mot: How can we hurdle frames to go beyond tracking-by-detection?'' \emph{IEEE Transactions on Image Processing}, pp. 4378--4392, 2023.

\bibitem{fu2022low}
Y.~Fu, Z.~Wang, T.~Zhang, and J.~Zhang, ``Low-light raw video denoising with a high-quality realistic motion dataset,'' \emph{IEEE Transactions on Multimedia}, pp. 8119--8131, 2022.

\bibitem{liu2022online}
Q.~Liu, D.~Chen, Q.~Chu, L.~Yuan, B.~Liu, L.~Zhang, and N.~Yu, ``Online multi-object tracking with unsupervised re-identification learning and occlusion estimation,'' \emph{Neurocomputing}, vol. 483, pp. 333--347, 2022.

\bibitem{wan2021tracking}
X.~Wan, J.~Cao, S.~Zhou, J.~Wang, and N.~Zheng, ``Tracking beyond detection: learning a global response map for end-to-end multi-object tracking,'' \emph{IEEE Transactions on Image Processing}, vol.~30, pp. 8222--8235, 2021.

\bibitem{liu2020gsm}
Q.~Liu, Q.~Chu, B.~Liu, and N.~Yu, ``Gsm: Graph similarity model for multi-object tracking.'' in \emph{International Joint Conference on Artificial Intelligence}, 2020, pp. 530--536.

\bibitem{li2023inference}
R.~Li, B.~Zhang, J.~Liu, W.~Liu, and Z.~Teng, ``Inference-domain network evolution: A new perspective for one-shot multi-object tracking,'' \emph{IEEE Transactions on Image Processing}, vol.~32, pp. 2147--2159, 2023.

\bibitem{vaswani2017attention}
A.~Vaswani, N.~Shazeer, N.~Parmar, J.~Uszkoreit, L.~Jones, A.~N. Gomez, {\L}.~Kaiser, and I.~Polosukhin, ``Attention is all you need,'' \emph{Advances in Neural Information Processing Systems}, vol.~30, 2017.

\bibitem{ren2015faster}
S.~Ren, K.~He, R.~Girshick, and J.~Sun, ``Faster r-cnn: Towards real-time object detection with region proposal networks,'' \emph{Advances in Neural Information Processing Systems}, vol.~28, 2015.

\bibitem{li2022grounded}
L.~H. Li, P.~Zhang, H.~Zhang, J.~Yang, C.~Li, Y.~Zhong, L.~Wang, L.~Yuan, L.~Zhang, J.-N. Hwang \emph{et~al.}, ``Grounded language-image pre-training,'' in \emph{Proceedings of the IEEE/CVF Conference on Computer Vision and Pattern Recognition}, 2022, pp. 10\,965--10\,975.

\bibitem{dai2021dynamic}
X.~Dai, Y.~Chen, B.~Xiao, D.~Chen, M.~Liu, L.~Yuan, and L.~Zhang, ``Dynamic head: Unifying object detection heads with attentions,'' in \emph{Proceedings of the IEEE/CVF Conference on Computer Vision and Pattern Recognition}, 2021, pp. 7373--7382.

\bibitem{gupta2019lvis}
A.~Gupta, P.~Dollar, and R.~Girshick, ``Lvis: A dataset for large vocabulary instance segmentation,'' in \emph{Proceedings of the IEEE/CVF Conference on Computer Vision and Pattern Recognition}, 2019, pp. 5356--5364.

\bibitem{shao2019objects365}
S.~Shao, Z.~Li, T.~Zhang, C.~Peng, G.~Yu, X.~Zhang, J.~Li, and J.~Sun, ``Objects365: A large-scale, high-quality dataset for object detection,'' in \emph{Proceedings of the IEEE/CVF International Conference on Computer Vision}, 2019, pp. 8430--8439.

\bibitem{carion2020end}
N.~Carion, F.~Massa, G.~Synnaeve, N.~Usunier, A.~Kirillov, and S.~Zagoruyko, ``End-to-end object detection with transformers,'' in \emph{Proceedings of the European Conference on Computer Vision}, 2020, pp. 213--229.

\bibitem{liu2021swin}
Z.~Liu, Y.~Lin, Y.~Cao, H.~Hu, Y.~Wei, Z.~Zhang, S.~Lin, and B.~Guo, ``Swin transformer: Hierarchical vision transformer using shifted windows,'' in \emph{Proceedings of the IEEE/CVF International Conference on Computer Vision}, 2021, pp. 10\,012--10\,022.

\bibitem{liu2024transformer}
Q.~Liu, Y.~Jiang, Z.~Tan, D.~Chen, Y.~Fu, Q.~Chu, G.~Hua, and N.~Yu, ``Transformer based pluralistic image completion with reduced information loss,'' \emph{IEEE Transactions on Pattern Analysis and Machine Intelligence}, 2024.

\bibitem{lai2024hyperspectral}
Z.~Lai, Y.~Fu, and J.~Zhang, ``Hyperspectral image super resolution with real unaligned rgb guidance,'' \emph{IEEE Transactions on Neural Networks and Learning Systems}, pp. 1--13, 2024.

\bibitem{jodoin2014urban}
J.-P. Jodoin, G.-A. Bilodeau, and N.~Saunier, ``Urban tracker: Multiple object tracking in urban mixed traffic,'' in \emph{Proceedings of the IEEE/CVF Winter Conference on Applications of Computer Vision}.\hskip 1em plus 0.5em minus 0.4em\relax IEEE, 2014, pp. 885--892.

\bibitem{bernardin2008evaluating}
K.~Bernardin and R.~Stiefelhagen, ``Evaluating multiple object tracking performance: the clear mot metrics,'' \emph{Journal on Image and Video Processing}, vol. 2008, pp. 1--10, 2008.

\bibitem{li2009learning}
Y.~Li, C.~Huang, and R.~Nevatia, ``Learning to associate: Hybridboosted multi-target tracker for crowded scene,'' in \emph{Proceedings of the IEEE/CVF Conference on Computer Vision and Pattern Recognition}, 2009, pp. 2953--2960.

\bibitem{couturier2021deep}
R.~Couturier, H.~N. Noura, O.~Salman, and A.~Sider, ``A deep learning object detection method for an efficient clusters initialization,'' \emph{arXiv preprint arXiv:2104.13634}, 2021.

\bibitem{redmon2018yolov3}
J.~Redmon and A.~Farhadi, ``Yolov3: An incremental improvement,'' \emph{arXiv preprint arXiv:1804.02767}, 2018.

\bibitem{he2015spatial}
K.~He, X.~Zhang, S.~Ren, and J.~Sun, ``Spatial pyramid pooling in deep convolutional networks for visual recognition,'' \emph{IEEE Transactions on Pattern Analysis and Machine Intelligence}, vol.~37, no.~9, pp. 1904--1916, 2015.

\bibitem{liu2018path}
S.~Liu, L.~Qi, H.~Qin, J.~Shi, and J.~Jia, ``Path aggregation network for instance segmentation,'' in \emph{Proceedings of the IEEE conference on computer vision and pattern recognition}, 2018, pp. 8759--8768.

\bibitem{he2016deep}
K.~He, X.~Zhang, S.~Ren, and J.~Sun, ``Deep residual learning for image recognition,'' in \emph{Proceedings of the IEEE/CVF Conference on Computer Vision and Pattern Recognition}, 2016, pp. 770--778.

\bibitem{krizhevsky2012imagenet}
A.~Krizhevsky, I.~Sutskever, and G.~E. Hinton, ``Imagenet classification with deep convolutional neural networks,'' \emph{Advances in Neural Information Processing Systems}, vol.~25, 2012.

\bibitem{meng2021conditional}
D.~Meng, X.~Chen, Z.~Fan, G.~Zeng, H.~Li, Y.~Yuan, L.~Sun, and J.~Wang, ``Conditional detr for fast training convergence,'' in \emph{Proceedings of the IEEE/CVF International Conference on Computer Vision}, 2021, pp. 3651--3660.

\bibitem{aharon2022bot}
N.~Aharon, R.~Orfaig, and B.-Z. Bobrovsky, ``Bot-sort: Robust associations multi-pedestrian tracking,'' \emph{arXiv preprint arXiv:2206.14651}, 2022.

\bibitem{loshchilov2018decoupled}
I.~Loshchilov and F.~Hutter, ``Decoupled weight decay regularization,'' in \emph{International Conference on Learning Representations}, 2018.

\bibitem{chen2021transformer}
X.~Chen, B.~Yan, J.~Zhu, D.~Wang, X.~Yang, and H.~Lu, ``Transformer tracking,'' in \emph{Proceedings of the IEEE/CVF Conference on Computer Vision and Pattern Recognition}, 2021, pp. 8126--8135.

\bibitem{wang2021transformer}
N.~Wang, W.~Zhou, J.~Wang, and H.~Li, ``Transformer meets tracker: Exploiting temporal context for robust visual tracking,'' in \emph{Proceedings of the IEEE/CVF Conference on Computer Vision and Pattern Recognition}, 2021, pp. 1571--1580.

\bibitem{meinhardt2022trackformer}
T.~Meinhardt, A.~Kirillov, L.~Leal-Taixe, and C.~Feichtenhofer, ``Trackformer: Multi-object tracking with transformers,'' in \emph{Proceedings of the IEEE/CVF Conference on Computer Vision and Pattern Recognition}, 2022, pp. 8844--8854.

\bibitem{milan2016mot16}
A.~Milan, L.~Leal-Taix{\'e}, I.~Reid, S.~Roth, and K.~Schindler, ``Mot16: A benchmark for multi-object tracking,'' \emph{arXiv preprint arXiv:1603.00831}, 2016.

\end{thebibliography}

\newpage

\vfill

\end{document}